\documentclass[a4paper,american,reqno]{amsart}

\usepackage[utf8]{inputenc}
\usepackage[T1]{fontenc}
\usepackage{babel}
\usepackage[binary-units=true]{siunitx}
\usepackage[algo2e,ruled,linesnumbered]{algorithm2e}
\usepackage{csquotes}
\usepackage{todonotes}
\usepackage{multicol}
\usepackage{comment}
\usepackage{booktabs}
\usepackage{flafter}
\usepackage{longtable}
\usepackage{dsfont}
\usepackage{placeins}
\usepackage{float}
\usepackage{amssymb}
\usepackage{subcaption}
\usepackage[font=small,labelfont=bf]{caption}
\usepackage{nomencl} 			
\usepackage{xcolor}				
\usepackage{mathtools}
\usepackage{multicol}
\usepackage{mathrsfs}
\usepackage{fancyvrb}
\usepackage{algorithm}
\usepackage{algpseudocode}
\usepackage{amsmath,amsthm,amssymb}
\usepackage{dsfont}
\usepackage{booktabs}

\usepackage{graphicx}			
\usepackage{microtype} 			

\definecolor{verde}{rgb}{0.2,0.6,0.2}
\definecolor{jpurple}{rgb}{0.9,0,0.9}
\definecolor{blue}{RGB}{41,5,195}

\usepackage{listings}

\usepackage[style = authoryear-comp,
            maxbibnames = 100,
            maxcitenames = 2,
            giveninits = true,
            uniquename = init,
            isbn = false,
            backend = bibtex]{biblatex}
\usepackage[colorlinks,
            citecolor=blue,
            urlcolor=blue,
            linkcolor=blue]{hyperref} 

\makeatletter
\patchcmd{\@settitle}{\uppercasenonmath\@title}{\scshape\large}{}{}
\patchcmd{\@setauthors}{\MakeUppercase}{\scshape\normalsize}{}{}
\makeatother

\vfuzz \hfuzz
\makeatletter
\@namedef{subjclassname@2020}{%
  \textup{2020} Mathematics Subject Classification}
\makeatother

\SetKwInOut{Input}{Input}
\SetKwInOut{Output}{Output}



\newcommand\MyBox[2]{
  \fbox{\lower0.75cm
    \vbox to 1.7cm{\vfil
      \hbox to 1.7cm{\hfil\parbox{1.4cm}{#1\\#2}\hfil}
      \vfil}%
  }%
}

\bibliography{FairML}

\begin{document}

\title[FairML: A \texttt{Julia} Package for Fair Classification]%
{FairML: A \texttt{Julia} Package for Fair Classification}

\author[J. P. Burgard, J. V. Pamplona]%
{Jan Pablo Burgard, João Vitor Pamplona}

\address[J. P. Burgard, J. V. Pamplona]{%
  Trier University,
  Department of Economic and Social Statistics,
  Universitätsring 15,
  54296 Trier,
  Germany}
\email{burgardj@uni-trier.de, pamplona@uni-trier.de}

\date{\today}

\begin{abstract}
In this paper, we propose \texttt{FairML.jl}, a \texttt{Julia} package providing a framework for fair classification in machine learning. In this framework, the fair learning process is divided into three stages. Each stage aims to reduce unfairness, such as disparate impact and disparate mistreatment, in the final prediction. For the preprocessing stage, we present a resampling method that addresses unfairness coming from data imbalances. 
The in-processing phase consist of a classification method. This can be either one coming from the \texttt{MLJ.jl} package, or a user defined one. For this phase, we incorporate fair ML methods that can handle unfairness to a certain degree through their optimization process. In the post-processing, we discuss the choice of the cut-off value for fair prediction. 
With simulations, we show the performance of the single phases and their combinations.
\end{abstract}

\keywords{Fair Machine Learning,
Optimization;
\texttt{Julia} language;
Mixed Models.
%
%
%
}

\maketitle
\markboth{Jan Pablo Burgard and João Vitor Pamplona}{FairML: A \texttt{Julia} Package for Fair Machine Learning}

\section{Introduction}
\label{sec:introduction}

The increase of automated decision-making necessitates the development of fair algorithms. These algorithms must adhere to societal values, particularly those that promote non-discrimination \parencite{importance}. While machine learning can offer precise classifications, depending on the data situation it can also inadvertently perpetuate classification biases in crucial domains like loan approvals \parencite{das2021fairness} and criminal justice \parencite{green2018fair}. For instance, loan approval algorithms may unfairly disadvantage single applicants by considering marital status. Similarly, criminal justice algorithms that associate race with recidivism risk can lead to discriminatory sentencing, neglecting individual circumstances. This underscores the critical need for fair classification frameworks to guarantee equal opportunities and outcomes, especially when applied within artificial intelligence application.

Driven by the growing concern of bias perpetuation by algorithms, the field of fair classification has seen a significant rise. Numerous research papers are now dedicated to exploring approaches that can mitigate bias across a wide range of algorithms. Notable examples include fair versions of logistic and linear regression \parencite{fairRL2}, support vector machine \parencite{FairSVM2}, random forests \parencite{fairRF}, decision trees \parencite{FairDT}, and generalized linear models (GLMs) \parencite{fairGLM}. These methods are designed to promote fair and equitable outcomes for all individuals by reducing potential biases that may stem from historical data or algorithmic design choices.

Moreover, in machine learning, training data for automated decision-making algorithms often originates from surveys. These surveys are usually designed using a sampling plan, which can deviate from the common assumption that each data point is sampled independently and with an equal probability of inclusion. Disregarding this can introduce additional bias. To mitigate this issue, approaches that handle mixed effects were proposed. Some examples can be seen in applications like Psychology \parencite{psicologia} and Medicine \parencite{medicina}. For a detailed discussion on survey methods and sampling strategies, see \textcite{Lohr2009}.

Packages for fair classification are already part of the literature, with versions available for \texttt{Python} \parencite{FAIRMPY} and \texttt{R} \parencite{FAIRMLR}. There also exists the \texttt{Fairness} package \parencite{fairnessJL} in \texttt{Julia}, aiming to equalize accuracies across sensitive groups. Although these packages present several techniques, none of them consider mixed effects. Moreover, the package developed for \texttt{R} considers the fairness metrics statistical parity, equality of opportunity and individual fairness. Our proposal focuses more on disparate impact, disparate mistreatment, false positive rate equality and false negative rate equality. We choose these metrics because they can be adapted as constraints in the model. Besides that, our package considers fairness as constraints, solving the constrained optimization problems via solver while the \texttt{Python} package uses other algorithms such as boosting tree \parencite{fairGBM} that penalizes unfairness. Additionally, our package handles mixed effects data.

The \texttt{Julia} programming language has been growing increasingly, especially in the field of machine learning. One reason is the availability of robust tools for optimization problems \parencite{JuliaBOM}. For this reason, a package for fair classification in \texttt{Julia} that takes into account an optimization problem adds value to the academic community.

This paper is organized as follows: In Section \ref{sec:chapter-1}, we establish the theoretical underpinnings of fair classification.
In Section \ref{sec:chapter-2}, we present a novel resampling method for preprocessing data with the aim of reducing disparate impact.
In Section \ref{sec:chapter-3}, we introduce optimization problems, previously proposed in the literature, that address unfairness metrics. There we also adapt the optimization methods for data with mixed effects.
In Section \ref{sec:chapter-4}, we present cross-validation-based post-processing methods to determine an optimal cut-off value for the classification process.
Finally,  in Section \ref{sec:numerical-results}, we conduct a comprehensive evaluation of our proposed package's effectiveness through various tests.
Our key findings and potential future directions are presented in Section \ref{sec:conclusion}.

\section{Machine Learning for fair classification}
\label{sec:chapter-1}

In machine learning, binary classification algorithms are used to estimate a specific classification $\hat{y}\in \{-1,1\}$ for a new data point $x$ based on a training set $\mathcal{D} = {(x^{\ell}, y_{\ell})}^n_{\ell=1}$, with $n$ being the number of points. For the point $x^\ell$ $\in$ $X = \left[x^1, \dotsc, x^n\right]$, if $y_\ell = 1$, we say that $x^\ell$ is in the positive class and if $y_\ell=-1$, $x^\ell$ belongs to the negative class for each $\ell \in [1,n] := \{1, \cdots, n\}$. Moreover,  $x^{\ell} \in \mathbb{R}^{p+1}$, for each $\ell \in [1,n]$, due to the addition of an extra column with the value $1$ as the data intercept.

When aiming for fairness in binary classification, we balance achieving good accuracy ($AC$) with ensuring fairness for observations $\ell$ based on their sensitive feature $s_\ell \in \{0, 1\}$, this is a standard approach in fair classification as stated by \textcite{pmlr-v54-zafar17a}. In this work the set of sensitive features is represented by $SF$, being the name of the sensitive variables.
While fairness in machine learning can be assessed through various metrics, in this paper we focus on disparate impact (DI) and disparate mistreatment (DM) that can be seen in \textcite{Paper1, bigdata}, respectively. The main reason is that they are already adapted to constraints within an optimization model, as demonstrated in \textcite{pmlr-v54-zafar17a}.

Considering the true labels and the predicted classifications of a supervised machine learning approach, we can categorize the data into four groups. A point is classified as true positive ($TP$) or true negative ($TN$) if its predicted class (positive or negative, respectively) matches its true label. Conversely, points are classified as false positives ($FP$) or false negatives ($FN$) if their predicted class differs from the true label. Based on this classification scheme, we can calculate accuracy, a metric where higher values indicate better classification performance. The formula for accuracy is as follows:
\begin{equation*}
    AC := \dfrac{TP + TN}{TP + TN + FP + FN} \in [0,1].
\end{equation*}
Now, we present the fairness metrics.
\subsection*{Disparate Impact}
Disparate impact refers to a situation where the probability under the prediction model $(\mathbb{P})$ is different conditional on  the sensitive feature values. A classifier is considered fair with respect to disparate impact if the probability of the point being classified as positive is equal when conditioning on the sensitive feature $s$, i.e.,
\begin{equation*}
\mathbb{P}(\hat{y}_\ell = 1 | s_\ell=0 ) = \mathbb{P}(\hat{y}_\ell = 1 | s_\ell=1 ).
\end{equation*}
To compute the disparate impact of a specific sensitive feature $s$ consider:
\begin{multicols}{2}
        \noindent
        \begin{align*}
        &\mathcal{S}_1 = \{x^\ell : \ell \in [1,n],\text{ } s_\ell = 1\}, \\
        &\mathcal{P} = \{x^\ell : \ell \in [1,n],\text{ } y_\ell = 1\}, \\
        &\mathcal{D}^{\mathcal{P}}_0 = \mathcal{S}_0 \cap \mathcal{P}, \\
        &\mathcal{D}^{\mathcal{P}}_1 = \mathcal{S}_1 \cap \mathcal{P},
        \end{align*}

        \noindent
        \begin{align}
        \begin{split}\label{eq4}
        &\mathcal{S}_0 = \{x^\ell : \ell \in [1,n],\text{ } s_\ell = 0\},\\
        &\mathcal{N} = \{x^\ell : \ell \in [1,n],\text{ } y_\ell = -1\}, \\
        &\mathcal{D}^{\mathcal{N}}_0 = \mathcal{S}_0 \cap \mathcal{N},\\
        &\mathcal{D}^{\mathcal{N}}_1 = \mathcal{S}_1 \cap \mathcal{N}.
        \end{split}
        \end{align}
\end{multicols}
Let $\mathcal{S}_0$ and $\mathcal{S}_1$ be disjoint subsets of dataset $X$, where the sensitive feature of all points in each subset is $0$ and $1$, respectively. Further, let $\mathcal{P}$ and $\mathcal{N}$ be the subsets where the true labels of the training set $\mathcal{D}$ are positive and negative, respectively. Then, we have the following metric di, based on \textcite{DIII}:
\begin{equation*}
    \text{di} := \dfrac{\vert\{\ell: \hat{y}_\ell = 1, x_\ell \in \mathcal{S}_0\}\vert}{\vert \mathcal{S}_0 \vert} \dfrac{\vert \mathcal{S}_1 \vert}{\vert\{\ell: \hat{y}_\ell = 1, x_\ell \in \mathcal{S}_1\}\vert} \in [0,\infty).
\end{equation*}
Note that di is the ratio between the proportion of points in $\mathcal{S}_0$ classified as positive and the proportion of points in $\mathcal{S}_1$ classified as positive. Hence disparate impact, as a metric, should ideally be equal to $1$ to indicate fair classifications. Values greater or lower than $1$ suggest the presence of unfairness. For instance, both $\text{di} = 2$ and $\text{di} = 0.5$ represent the same amount of discrimination, but in opposite directions. To address this limitation and achieve a more nuanced metric, we use the minimum value between $\text{di}$ and its inverse $\tfrac{1}{\text{di}}$. Furthermore, to align with the convention of other fairness metrics where a value closer to $0$ indicates greater fairness (as will be show later), we redefine the \text{DI} as follows

\begin{equation}\label{DI}
    \text{DI} := 1 - \min (\text{di}, \text{di}^{-1})  \in [0,1].
\end{equation}
Hence, a value closer to $0$ indicates better performance and a value closer to $1$ indicates worse performance.
 
\subsection*{Disparate Mistreatment}
Disparate mistreatment, also known as equalized odds \parencite{EQOODS}, is defined as the condition in which the misclassification rates for points with different values in the sensitive features are unequal. In other words, a classification is free of disparate mistreatment when the classification algorithm is equally likely to misclassify points in both positive and negative classes, regardless of their sensitive characteristics.

A classification is considered free of disparate mistreatment if the rate of false positives and false negatives is equal for both categories of a sensitive feature $s$. That is, 
\begin{equation*}
\mathbb{P}(\hat{y}_\ell = 1 | \ell \in \mathcal{D}^{\mathcal{N}}_0) = \mathbb{P}(\hat{y}_\ell = 1 | \ell \in \mathcal{D}^{\mathcal{N}}_1)
\end{equation*}
and
\begin{equation*}
\mathbb{P}(\hat{y}_\ell = -1 | \ell \in \mathcal{D}^{\mathcal{P}}_0) = \mathbb{P}(\hat{y}_\ell = -1 | \ell \in \mathcal{D}^{\mathcal{P}}_1).
\end{equation*}
To quantify the disparate mistreatment with respect to a specific sensitive feature $s$, we first establish the equations for the false positive rate ($FPR$) and false negative rate ($FNR$) metrics. The $FPR$ metric is defined as the absolute value of the difference between the false positive rates of the categories defined by the sensitive feature $s$, as follows:
\begin{align*}
    FPR &:= \vert FPR_{s=0} - FPR_{s=1} \vert\\
    &= \Big\vert \dfrac{FP_{s=0}}{FP_{s=0} + TN_{s=0}} - \dfrac{FP_{s=1}}{FP_{s=1} + TN_{s=1}}  \Big\vert  \label{FPR}\tag{FPR} \\
    &= \Big\vert \dfrac{\vert\{\ell: \hat{y}_\ell = 1, x^\ell \in \mathcal{D}^{\mathcal{N}}_0\}\vert}{\vert \mathcal{D}^{\mathcal{N}}_0 \vert}  - \dfrac{\vert\{\ell: \hat{y}_\ell = 1, x^\ell \in \mathcal{D}^{\mathcal{N}}_1\vert}{\vert \mathcal{D}^{\mathcal{N}}_1 \vert}  \Big\vert \in [0,1].
\end{align*}
Similarly, the $FNR$ is given by:
\begin{align*}
    FNR &:= \vert FNR_{s=0} - FNR_{s=1} \vert \\
    &= \Big\vert \dfrac{FN_{s=0}}{FN_{s=0} + TP_{s=0}} - \dfrac{FN_{s=1}}{FN_{s=1} + TP_{s=1}}  \Big\vert  \label{FNR}\tag{FNR} \\
    &= \Big\vert \dfrac{\vert\{\ell: \hat{y}_\ell = -1, x^\ell \in \mathcal{D}^{\mathcal{P}}_0\}\vert}{\vert \mathcal{D}^{\mathcal{P}}_0 \vert}  - \dfrac{\vert\{\ell: \hat{y}_\ell = -1, x^\ell \in \mathcal{D}^{\mathcal{P}}_1\}\vert}{\vert \mathcal{D}^{\mathcal{P}}_1 \vert} \Big\vert  \in [0,1],
\end{align*}
Disparate mistreatment is the mean of both metrics above. Again, the lower  the value, the fairer classification. 
\begin{equation}\label{DM}\tag{DM}
    DM = \dfrac{FPR + FNR}{2}  \in [0,1].
\end{equation}

With our fairness metrics at hand, we now present the strategy of our \texttt{Julia} package, \texttt{FairML} that employs a variety of optimization techniques and a resampling strategy to ensure fairness in classifications based on a user-specified sensitive attribute. The package operates under a three-step framework:
\begin{enumerate}
    \item Preprocessing: This stage encompasses the implementation of functions that perform initial data manipulation aimed at enhancing fairness metrics;
    
    \item In-processing:  This stage constitutes the main part of the paper, where optimization problems are addressed with the aim of improving a specific fairness metric;
    
    \item Post-processing: Following the previous stage, which outputs class membership probabilities, this phase is responsible for performing classification. It may or may not employ strategies to optimize a specific fairness metric in relation to accuracy.
\end{enumerate}

While the theoretical underpinnings, construction, and explanation of each stage will be detailed in subsequent chapters, the package's core functionality unifies all stages into a single, user-friendly interface:

\begin{figure}[H]
\centering
\includegraphics[width=12.65cm]{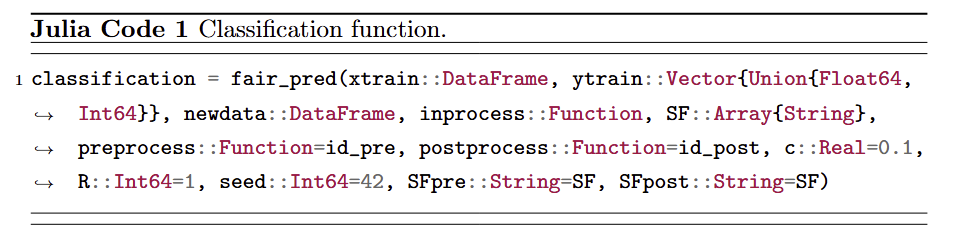}
\end{figure}


Besides that, many datasets exhibit unexplained variation within groups or across different levels, more details can be seen in Section \ref{sec:chapter-3}. Hence, in this package we also propose a classification function for this type of data: 

\begin{figure}[H]
\centering
\includegraphics[width=12.65cm]{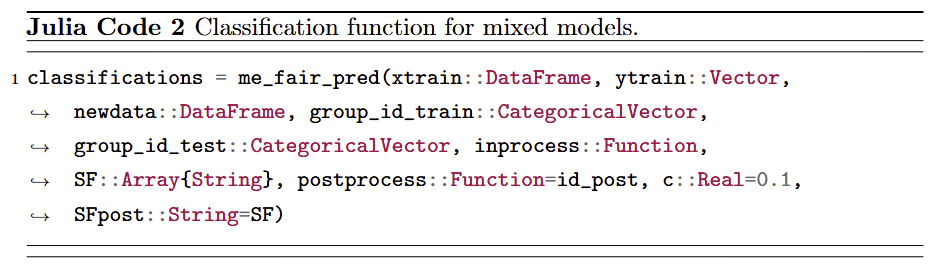}
\end{figure}
Being:
\begin{itemize}
    \item Input arguments:
            \begin{enumerate}
                \item $xtrain$: The dataset that the labels are known (training set);
                \item $ytrain$: The labels of the dataset $xtrain$;
                \item $newdata$: The new dataset for which we want to obtain the $classifications$;
                \item $inprocess$: One of the several optimization problems available in this package or any machine learning classification method present in MLJ.jl package;
                \item $SF$: One or a set of sensitive features (variables names. E.g Sex, race\dots), that will act in the in-processing phase. If the algorithm come from the MLJ.jl package, no fair constraint are acting in this phase;
                \item $group\_id\_train$: Training set group category;
                \item $group\_id\_newdata$: New dataset group category.
            \end{enumerate}

    \item Optional argument:
            \begin{enumerate}
                \item $preprocess$: A pre-processing function among the options available in this package, $id\_pre()$ by default;
                \item $postprocess$: A post-processing function among the options available in this package, $id\_post()$ by default;
                \item $c$: The threshold of the fair optimization problems, $0.1$ by default;
                \item $R$: Number of iterations of the preprocessing phase, each time sampling differently using the resampling method, $1$ by default;
                \item $seed$: For sample selection in $R$, $42$ by default;
                \item $SFpre$: One sensitive features (variable name), that will act in the preprocessing phase, disabled by default;
                \item $SFpost$: One sensitive features (variable name), that will act in the post-processing phase, disabled by default.
            \end{enumerate}

    \item Output arguments:
             \begin{enumerate}
                \item $classifications$: Classifications of the $newdata$ points.
            \end{enumerate}

\end{itemize}

The classification function for mixed models ignores the preprocessing phase, as this phase tends to eliminate numerous data points, as discussed in \ref{sec:chapter-2}. Such elimination can lead to empty groups, which is not permissible in the classification functions for mixed models. 

It is essential to highlight that both the preprocessing and post-processing stages should be limited to handling a single sensitive feature each. Only the in-processing stage can handle with multiple sensitive features at the same time, creating multiples fairness constraints for the optimization problems. However, sensitive features can differ across the three phases with the aim to achieve fairness through various potential discrimination classes.
\section{Preprocessing}
\label{sec:chapter-2}

Resampling methods can serve various purposes, as can be seen in \textcite{resamp}. In our case, the goal is to mitigate disparate impact or disparate mistreatment in the data. We achieve this by generating multiple datasets that exhibit less unfairness than the original. In this context, we developed a hybrid approach that combines an adapted undersampling technique with cross-validation to address this issue.

Undersampling \parencite{mohammed2020machine} reduces the majority class, in the sensitive feature, to balance the dataset, while cross-validation \parencite{blagus2015joint} provides a evaluation of the model by iteratively training and testing on different subsets. Similar approaches have been used for class-imbalanced data in \textcite{zughrat2014support} and \textcite{FAIRMPY}.

As indicated by Equation \eqref{DI}, regarding to disparate impact, our goal is to ensure equal representation of positive and negative labels across both categories of the sensitive features. To achieve this, we enforce this condition within the training set $\mathcal{D}$ using the following strategy:

\begin{enumerate}
    \item Separate the training data $\mathcal{D}$ as in Equation \eqref{eq4};

     \item Compute the size of the smallest among the four subsets:
        \[  
        J = \min(\vert \mathcal{D}^{\mathcal{N}}_0\vert, \vert\mathcal{D}^{\mathcal{N}}_1\vert, \vert\mathcal{D}^{\mathcal{P}}_0\vert, \vert\mathcal{D}^{\mathcal{P}}_1\vert).
        \]

    \item  For each subset do a random sampling with replacement, $M$ as follows:

    \begin{multicols}{2}
    \noindent
        \[
        M_J^{\mathcal{D}^{\mathcal{N}}_0} \subseteq \mathcal{D}^{\mathcal{N}}_0 \text{, with } \vert M_J^{\mathcal{D}^{\mathcal{N}}_0} \vert = J,
        \]
        \[
        M_J^{\mathcal{D}^{\mathcal{N}}_1} \subseteq \mathcal{D}^{\mathcal{N}}_1 \text{, with } \vert M_J^{\mathcal{D}^{\mathcal{N}}_1} \vert = J,
        \]

        \noindent
        \[
        M_J^{\mathcal{D}^{\mathcal{P}}_0} \subseteq \mathcal{D}^{\mathcal{P}}_0 \text{, with } \vert M_J^{\mathcal{D}^{\mathcal{P}}_0} \vert = J,
        \]
        \[
        M_J^{\mathcal{D}^{\mathcal{P}}_1} \subseteq \mathcal{D}^{\mathcal{P}}_1 \text{, with } \vert M_J^{\mathcal{D}^{\mathcal{P}}_1} \vert = J.
        \]
        \end{multicols}
    
    \item Create the new training dataset:
    \[
    \mathcal{D} = M_J^{\mathcal{D}^{\mathcal{N}}_0} \cup M_J^{\mathcal{D}^{\mathcal{N}}_1} \cup M_J^{\mathcal{D}^{\mathcal{P}}_0} \cup M_J^{\mathcal{D}^{\mathcal{P}}_1}.
    \]
    
\end{enumerate}
Therefore, since there is no disproportionality of labels across different sensitive features categories, we expected to have a new dataset with less disparate impact than the previous one.

Observe that the generation of the new dataset is a random process. To account for the insecurity introduced by the random generation, we allow the user to define the number $R$ of times this data set is to be generated. In the pre-processing phase, the  best one is chosen as follows:
\begin{enumerate}
    \item Do the preprocessing phase $R$ times, generating $R$ different datasets;
     \item For each dataset:
     \begin{enumerate}
         \item Calculate the coefficients using the in-processing phase;
         \item Compute the classifications on the full training set (before resampling);
         \item Use the classifications to calculate disparate impact or disparate mistreatment; 
     \end{enumerate}
    \item Select the classification with the best fairness metric value;
    \item Use the coefficients from the best classification to calculate classifications on new data.    
\end{enumerate}

That is, from all the $R$ calculated coefficients, this phase selects the one that generate the smallest disparate impact or disparate mistreatment on the full training set, and uses it to classify the points in the new dataset (input $newdata$).

While the algorithm was designed to address disparate impact, preliminary numerical tests have shown that it can also mitigate disparate treatment. This makes it a flexible tool, allowing the user to choose the specific focus.

The inputs and outputs of the preprocessing function (di\_pre) are documented on the \href{https://github.com/JoaoVitorPamplona/FairML.jl}{package's GitHub page}.

In the next section, we will explain the in-processing phase.

\section{In-processing}
\label{sec:chapter-3}

The main goal of the in-processing phase is to predict the probability of a new point being classified as $1$ or $-1$. This is achieved by finding the coefficients of a prediction model by solving an optimization problem. We propose several optimization problems that can improve the fairness metrics of disparate impact, false positive rate, false negative rate, and disparate mistreatment.

This paper mainly focuses on two methods for binary classification. The first approach is logistic regression (LR). Since in our data we have $y \in \{-1,1\}$, we adapt, w.l.o.g., the logistic regression model \parencite{john2004mp}.

\begin{align}\label{LR}\tag{LR}
\begin{split}
\displaystyle \min_{\beta}\hspace{0.1cm} & -\displaystyle\sum_{\ell=1}^{n} \Big[\Big(\frac{y_{\ell} + 1}{2}\Big) \log(m_{\beta}^{LR}(x^{\ell})) + \Big(\frac{y_{\ell} - 1}{2}\Big)\log(1 - m_{\beta}^{LR}(x^{\ell}))\Big]
\end{split}
\end{align}
with the prediction function given by
\begin{equation}\label{PREDEQLR}
     m_{\beta}^{LR}(x) :=  \dfrac{1}{1 + e^{-\beta^\top x}}.
\end{equation}

The second method is the standard Support Vector Machine (SVM), proposed by \textcite{vapnik1964class} and \textcite{hearst1998support}.
\begin{align}\label{SVM}\tag{SVM}
\begin{split}
\underset {\left(\beta, \xi\right)}{\min} &\;\;\;\;
\frac{1}{2} \Vert \beta \Vert^2 + \mu \sum_{\ell=1}^n \xi_{\ell}\\ 
\rm{s.t} &\;\;\;\; y_{\ell}(m_{\beta}^{SVM}(x_\ell)) \geq 1 - \xi_{\ell}, \text{  } \ell = 1, \dots, n
\end{split}
\end{align}
with the prediction function given by
\begin{equation}\label{PREDEQSVM}
     m_{\beta}^{SVM}(x) :=  \beta^\top x.
\end{equation}

As already mentioned, the first column of the matrix $X$ should be a vector of ones, that is, the first entrance of $x^{\ell}, \forall \ell \in [1,n],$ is equal to $1$. If this column does not exist, the functions of this package automatically add one. Note that in standard SVM implementations, an intercept term is typically not added to the data, but rather a so called bias is included in the problem constraints. Using the formulation of \textcite{hsieh2008dual}, we can adjust it to include an intercept term being the first entry in $\beta$.

In problems \eqref{SVM} and \eqref{LR}, fairness constraints can be added. Let us now present them, based on the formulations of \textcite{pmlr-v54-zafar17a}.

\subsubsection*{Fairness Constraints for Disparate Impact}
As stated in Expression \eqref{DI}, to ensure a classification is free from disparate impact, the conditional probabilities of a positive classification given the sensitive feature s should be equal. While achieving zero disparate impact is a desirable goal, it can potentially reduce the classification accuracy, as we have a trade-off between fairness and accuracy \parencite{tradeoff1,tradeoff2}. To address this trade-off, \textcite{pmlr-v54-zafar17a} suggest introduce a fairness threshold, denoted by $c \in \mathbb{R}^+$, which allows us to adjust the relative importance placed on fairness compared to accuracy. With this logic, we deduce the following constraints:

\begin{align}
\begin{split}
\hspace{0.1cm} & \frac{1}{n}\displaystyle\sum_{\ell=1}^{n} \displaystyle (s_{\ell} - \bar{s})(\beta^\top x^{\ell}) \leq c\\
              & \frac{1}{n}\displaystyle\sum_{\ell=1}^{n} (s_{\ell} - \bar{s})(\beta^\top x^{\ell}) \geq - c\label{ret1}.
\end{split}
\end{align}

A more detailed description of how disparate impact constraints are constructed is provided in \textcite{pamplona2024fairSVM}. Note that these constraints take into account the inner product $\beta^\top x^{\ell}$, which is the main component in both prediction functions \eqref{PREDEQSVM} and \eqref{PREDEQLR}.

\subsubsection*{Fairness Constraints for Disparate Mistreatment}
As previously discussed, in Section \ref{sec:chapter-1}, the fairness metric disparate mistreatment aims to simultaneously equalize or approximate (depending on $c$) the false negative rate and false positive rate across the different categories of the sensitive feature.

We begin by considering the $FNR$ constraint. A point is a false negative if $y_\ell = 1$ and $\beta^{\top} x^\ell < 0$, that is, if and only if
\begin{equation}\label{zaf1}
    \min(0, \frac{1+y_\ell}{2} y_\ell \beta^{\top} x^\ell)
\end{equation}
is greater than zero, being $\beta$ the coefficient. In fact, let us examine all four possibilities:
\begin{enumerate}
        \item True Negative: $y_\ell = -1$ and $\beta^{\top} x^\ell < 0$ $\implies$ $\min(0, \frac{1+y_\ell}{2} y_\ell \beta^{\top} x^\ell) = 0$
        
        \item False Positive: $y_\ell = -1$ and $\beta^{\top} x^\ell > 0$ $\implies$ $\min(0, \frac{1+y_\ell}{2} y_\ell \beta^{\top} x^\ell) = 0$ 
        
        \item False Negative: $y_\ell = 1$ and $\beta^{\top} x^\ell < 0$ $\implies$ $\min(0, \frac{1+y_\ell}{2} y_\ell \beta^{\top} x^\ell) = \beta^{\top} x^\ell$ 
        
        \item True Positive: $y_\ell = 1$ and $\beta^{\top} x^\ell > 0$ $\implies$ $\min(0, \frac{1+y_\ell}{2} y_\ell \beta^{\top} x^\ell) = 0$  
\end{enumerate}

For this reason, \textcite{DISPMIST} uses the Expression \eqref{zaf1} to select the false negative points among all points. However, note that in the $FNR$ constraint, we only need to care about the points that belong to $\mathcal{P}$, because for the point that belongs to $\mathcal{N}$ the Expression \eqref{zaf1} is always equal to $0$. Since for a point $x^{\ell} \in \mathcal{P}$ we have $y_\ell = 1$, Expression \eqref{zaf1} becomes $\min(0,\beta^\top x^{\ell})$.

To obtain the same proportion of false negatives in both sensitive categories, the $FNR$ constraints impose that the sums of the minimum between $0$ and the inner products of the coefficient and a positive point are close to each other in each sensitive category, as follows:

\begin{subequations}\label{ret9.0}
\begin{align}
\hspace{0.1cm} & \frac{\vert \mathcal{S}_0 \vert}{n}\displaystyle\sum_{x^\ell \in \mathcal{D}^{\mathcal{P}}_1} \min(0,\beta^\top x^{\ell}) - \frac{\vert \mathcal{S}_1 \vert}{n}\displaystyle\sum_{x^\ell \in \mathcal{D}^{\mathcal{P}}_0} \min(0,\beta^\top x^{\ell}) \leq c\label{ret9.1} \\ 
             &\frac{\vert \mathcal{S}_0 \vert}{n}\displaystyle\sum_{x^\ell \in \mathcal{D}^{\mathcal{P}}_1} \min(0,\beta^\top x^{\ell}) - \frac{\vert \mathcal{S}_1 \vert}{n}\displaystyle\sum_{x^\ell \in \mathcal{D}^{\mathcal{P}}_0} \min(0,\beta^\top x^{\ell}) \geq -c\label{ret9.2} 
\end{align}
\end{subequations}
For false positive points, we employ the same logic, however, replacing the Expression \eqref{zaf1} with:
\begin{equation*}
    \min(0, \frac{1-y_\ell}{2} y_\ell \beta^{\top} x^\ell),
\end{equation*}
and hence
\begin{enumerate}
 
        \item True Negative: $y_\ell = -1$ and $\beta^{\top} x^\ell < 0$ $\implies$ $\min(0, \frac{1-y_\ell}{2} y_\ell \beta^{\top} x^\ell) = 0$
        
        \item False Positive: $y_\ell = -1$ and $\beta^{\top} x^\ell > 0$ $\implies$ $\min(0, \frac{1-y_\ell}{2} y_\ell \beta^{\top} x^\ell) = -\beta^{\top} x^\ell$ 
        
        \item False Negative: $y_\ell = 1$ and $\beta^{\top} x^\ell < 0$ $\implies$ $\min(0, \frac{1-y_\ell}{2} y_\ell \beta^{\top} x^\ell) = 0$ 
        
        \item True Positive: $y_\ell = 1$ and $\beta^{\top} x^\ell > 0$ $\implies$ $\min(0, \frac{1-y_\ell}{2} y_\ell \beta^{\top} x^\ell) = 0$ 
 
\end{enumerate}
That is, in the $FPR$ constraints, we only need to care about the points that belong to $\mathcal{N}$. Similarly to the $FNR$ constraints, the $FPR$ constraints impose that the sums of the minimum between $0$ and minus the inner products of the coefficient and a negative point are close to each other in each sensitive category. That is,  

\begin{subequations}\label{ret9.01}
\begin{align}
\hspace{0.1cm}
             & \frac{\vert \mathcal{S}_0 \vert}{n}\displaystyle\sum_{x^\ell \in \mathcal{D}^{\mathcal{N}}_1} \min(0,-\beta^\top x^{\ell}) - \frac{\vert \mathcal{S}_1 \vert}{n}\displaystyle\sum_{x^\ell \in \mathcal{D}^{\mathcal{N}}_0} \min(0,-\beta^\top x^{\ell}) \leq c\label{ret9.3} \\ 
              &\frac{\vert \mathcal{S}_0 \vert}{n}\displaystyle\sum_{x^\ell \in \mathcal{D}^{\mathcal{N}}_1} \min(0,-\beta^\top x^{\ell}) - \frac{\vert \mathcal{S}_1 \vert}{n}\displaystyle\sum_{x^\ell \in \mathcal{D}^{\mathcal{N}}_0} \min(0,-\beta^\top x^{\ell}) \geq -c \label{ret9.4}
\end{align}
\end{subequations}

Therefore, the Disparate Mistreatment constraints are a combination of Constraints~\eqref{ret9.1},\eqref{ret9.2},\eqref{ret9.3} and \eqref{ret9.4}.

Given the constraints we have presented, we can utilize the following problems in the in-processing phase:

\begin{itemize}
    \item Logistic regression free of disparate impact;
    \item Logistic regression free of false negative rate;
    \item Logistic regression free of false positive rate;
    \item Logistic regression free of disparate mistreatment;
    
    \item Support vector machine free of disparate impact;
    \item Support vector machine free of false negative rate;
    \item Support vector machine free of false positive rate;
    \item Support vector machine free of disparate mistreatment.
\end{itemize}

Problems \eqref{LR}, \eqref{SVM} and above do not deal with random effects, which can be happening in diverse application, like from medicine or psychology \parencite{psicologia,medicina}. However, these problems, like many other statistical models, can lead to unfair outcomes. In light of this, we propose a novel research area designated as fair machine classification for data with mixed effects \parencite{pamplona2024fairSVM, burgard2024fairLR}. We adapt well-established methods from the literature to address fair machine learning optimization problems in the presence of random effects.

\subsection*{Mixed Model}
To capture the latent heterogeneity present in some types of data, which can encompasses cultural, demographic, biological, and behavioral aspects, it is imperative to incorporate random effects into the predictive model. Omitting these effects can lead to substantial bias in the classifications, compromising the accuracy and generalization of the results \parencite{heter1, heter2}.

Let $g$ being the random vector and $g_i$ with $i ~\in~ [1,K]$, representing the group-specific random effect, with $g$ following a normal distribution with mean zero. Consider $\Gamma_i$ the size of the group $i$ for each $i ~\in~ [1,K]$ and $y_{ij}$ the label of $(x^{ij})^\top~=~(x^{ij}_1, \dots, x^{ij}_p)$ with $j \in [1,\Gamma_i]$. 

To ensure that in all of our problems we have $y \in \{-1,1\}$, we adapt, w.l.o.g., the mixed effects logistic regression model as we did in \eqref{LR}.
\begin{align}\label{MELR}\tag{MELR}
\displaystyle \min_{\beta,g}\hspace{0.1cm} & -\displaystyle\sum_{i=1}^K \displaystyle\sum_{j=1}^{\Gamma_i} \Big[\Big(\frac{y_{ij} + 1}{2}\Big)  \log(m_{\beta,g}^{LR}(x^{ij})) + \Big(\frac{y_{ij} - 1}{2}\Big)\log(1 - m_{\beta,g}^{LR}(x^{ij}))\Big] + \lambda \sum_{i=1}^K g_i^2
\end{align}
with the prediction function given by
\begin{equation}\label{PREDEQMELR}
    m_{\beta,g}^{LR}(x^{ij}) := \dfrac{1}{1 + e^{-(\beta^\top x^{ij} + g_i)}},
\end{equation}
and $y_{ij}$ being the label in the observation $j$ in group $i$ and $j~\in~[1, \Gamma_i]$, and $\Gamma_i$ the size of the group $i$. For a detailed explanation and a breakdown of the Mixed Effects Logistic Regression derivation, see \textcite{burgard2024fairLR}. For the Mixed Effects Support Vector Machine, we consider the model present by \textcite{pamplona2024fairSVM}:

\begin{align}\label{MESVM}\tag{MESVM}
\begin{split}
\underset {\left(\beta, g, \xi \right )}{\min} &\;\;\;\;
\frac{1}{2} \Vert \beta \Vert^2 + \mu \sum_{i=1}^K \sum_{j=1}^{\Gamma_i} \xi_{ij} + \lambda \sum_{i=1}^K g_i^2\\ 
\rm{s.t} &\;\;\;\; y_{ij}(m_{\beta,g}^{SVM}(x^{ij})) \geq 1 - \xi_{ij}, \text{  } i = [1, K], \text{  } j = [1, \Gamma_i]
\end{split}
\end{align}
with the prediction function given by
\begin{equation}\label{PREDEQMESVM}
    m_{\beta,g}^{SVM}(x^{ij}) := \beta^\top x_{ij} + g_i.
\end{equation}

In mixed models, all constraints previously constructed for regular models are adapted to account for the existence of the random effect. The construction logic for these constraints is equivalent to the problems with only fixed effects, with an adaptation of the created subgroups as shown in \eqref{eq4} as follows:
\begin{multicols}{2}
        \noindent
        \begin{align*}
        &\mathcal{S}_1^i = \{x^{ij} : j \in [1,\Gamma_i],\text{ } s_{ij} = 1\}, \\
        &\mathcal{P}^i = \{x^{ij} : j \in [1,\Gamma_i],\text{ } y_{ij} = 1\} \\
        &\mathcal{D}^{\mathcal{P}^i}_0 = \mathcal{S}_0^i \cap \mathcal{P}^i, \\
        &\mathcal{D}^{\mathcal{P}^i}_1 = \mathcal{S}_1^i \cap \mathcal{P}^i,
        \end{align*}

        \noindent
        \begin{align*}
        &\mathcal{S}_0^i = \{x^{ij} : j \in [1,\Gamma_i],\text{ } s_{ij} = 0\},\\
        &\mathcal{N}^i = \{x^{ij} : j \in [1,\Gamma_i],\text{ } y_{ij} = -1\} , \\
        &\mathcal{D}^{\mathcal{N}^i}_0 = \mathcal{S}_0^i \cap \mathcal{N}^i,\\
        &\mathcal{D}^{\mathcal{N}^i}_1 = \mathcal{S}_1^i \cap \mathcal{N}^i.
        \end{align*}
\end{multicols}
Observe that each subset is created for each cluster $i \in [1,K]$.

Moreover, we need to modify the fairness constraints to account for random effects. 
\subsubsection*{Disparate Impact}
Following the same logic as presented before, but considering a group-to-group analysis, we have a similar construction for the disparate impact constraints in mixed models that can be seen in \textcite{pamplona2024fairSVM} and is given by:

\begin{align*}
\begin{split}
\hspace{0.1cm} & \frac{1}{n}\displaystyle\sum_{i=1}^K \displaystyle\sum_{j=1}^{\Gamma_i} (s_{ij} - \bar{s})(\beta^\top x^{ij} + g_i) \leq c,\\
              & \frac{1}{n}\displaystyle\sum_{i=1}^K \displaystyle\sum_{j=1}^{\Gamma_i} (s_{ij} - \bar{s})(\beta^\top x^{ij} + g_i) \geq -c. 
\end{split}
\end{align*}

\subsubsection*{Disparate Mistreatment}
We now discuss the $\text{DM}$ metric for mixed effects. For the $FNR$ constraints, we adapt the Expression \eqref{zaf1} to incorporate the random effects as follows:
\begin{equation*}
    \min\Big(0, \frac{1+y_{ij}}{2} y_{ij}(\beta^{\top} x^{ij} + g_i)\Big).
\end{equation*}

As done for the regular models, we only need take care about the positive points. And, for these points, the expression above becomes $\min(0,\beta^\top x^{ij} + g_i)$.

On the other hand, for the $FPR$ constraints, the selection of the false positive points is adapted to
\begin{equation*}
    \min\Big(0, \frac{1-y_{ij}}{2} y_{ij} (\beta^{\top} x^{ij} + g_i)\Big).
\end{equation*}
Here we only need to take care about the negative points. And, for these points, the expression above becomes $\min(0,-\beta^\top x^{ij} - g_i)$. Combining all constraints yields the following set of constraints for a classification free of disparate mistreatment in mixed models:

\begin{align*}
\begin{split}
\hspace{0.1cm} & \frac{\vert \mathcal{S}_0 \vert}{n}\displaystyle\sum_{i=1}^K \displaystyle\sum_{x^{ij} \in \mathcal{D}^{\mathcal{P}^i}_1} \min(0,\beta^\top x^{ij} + g_i) - \frac{\vert \mathcal{S}_1 \vert}{n}\displaystyle\sum_{i=1}^K \displaystyle\sum_{x^{ij} \in \mathcal{D}^{\mathcal{P}^i}_0} \min(0,\beta^\top x^{ij} + g_i) \leq c\\
& \frac{\vert \mathcal{S}_0 \vert}{n}\displaystyle\sum_{i=1}^K \displaystyle\sum_{x^{ij} \in \mathcal{D}^{\mathcal{P}^i}_1} \min(0,\beta^\top x^{ij} + g_i) - \frac{\vert \mathcal{S}_1 \vert}{n}\displaystyle\sum_{i=1}^K \displaystyle\sum_{x^{ij} \in \mathcal{D}^{\mathcal{P}^i}_0} \min(0,\beta^\top x^{ij} + g_i) \geq -c\\
& \frac{\vert \mathcal{S}_0 \vert}{n}\displaystyle\sum_{i=1}^K \displaystyle\sum_{x^{ij} \in \mathcal{D}^{\mathcal{N}^i}_1} \min(0,-\beta^\top x^{ij} - g_i) - \frac{\vert \mathcal{S}_1 \vert}{n}\displaystyle\sum_{i=1}^K \displaystyle\sum_{x^{ij} \in \mathcal{D}^{\mathcal{N}^i}_0} \min(0,-\beta^\top x^{ij} - g_i) \leq c\\
& \frac{\vert \mathcal{S}_0 \vert}{n}\displaystyle\sum_{i=1}^K \displaystyle\sum_{x^{ij} \in \mathcal{D}^{\mathcal{N}^i}_1} \min(0,-\beta^\top x^{ij} - g_i) - \frac{\vert \mathcal{S}_1 \vert}{n}\displaystyle\sum_{i=1}^K \displaystyle\sum_{x^{ij} \in \mathcal{D}^{\mathcal{N}^i}_0} \min(0,-\beta^\top x^{ij} - g_i) \geq -c.
\end{split}
\end{align*}
\normalsize

The first summation iterates over all groups, while the second summation iterates only over the desired points within each group. 

Similarly to regular models, we can assign the constraints above to problems \eqref{MELR} and \eqref{MESVM}, leading to $8$ new additional optimization problems, which are:

\begin{itemize}
    \item Mixed effects logistic regression free of disparate impact
    \item Mixed effects logistic regression free of false negative rate
    \item Mixed effects logistic regression free of false positive rate
    \item Mixed effects logistic regression free of disparate mistreatment
    
    \item Mixed effects support vector machine free of disparate impact
    \item Mixed effects support vector machine free of false negative rate
    \item Mixed effects support vector machine free of false positive rate
    \item Mixed effects support vector machine free of disparate mistreatment
\end{itemize}

Unlike regular models, mixed model algorithms cannot be replaced by \texttt{MLJ} models, as the latter are not suitable for this kind of problem. 

It is worth to remember that all constraints, both for the regular model and the model that includes random effects, allow for the use of multiple sensitive features simultaneously.

The inputs and outputs of all in-processing functions are documented on the \href{https://github.com/JoaoVitorPamplona/FairML.jl}{package's GitHub page} and in the next section, we will explain the post-processing phase.
\section{Post-processing}
\label{sec:chapter-4}

The post-processing phase implements an algorithm that seeks an optimal cut-off value for classification \parencite{ren2016prospective, cheong2013optimal}. An approach that implements a similar strategy, but considering each sensitive group, can be seen in \textcite{FAIRMPY}. In our approach, we consider the entire dataset to ensure that no particular sensitive group is at advantaged or disadvantaged.

Classifications are computed using the predicted probability values from both the training and testing sets obtained from the previous phase. 

Given the predicted probabilities from both the training and new datasets obtained in the in-processing phase:

\begin{enumerate}
    \item For each cut-off value $v$ ranging from $0.01$ to $0.99$ (with an increment of $0.01$), do:

    \begin{itemize}
        \item Generate classifications for training set as follows: if the probability is greater or equal $v$, classify as positive, otherwise as negative;

        \item Compute the accuracy ($AC_v$) and the desired fairness metric value ($fm_v$) for training set.
    \end{itemize}

    \item Select only the values of $v$ that decrease at most 5\% of the accuracy compared to the accuracy given by the cut-off value $v = 0.5$.  Among them, select the best result using $B = argmax_v(AC_v-fm_v)$;
    
    \item Use the new cut-off value, $B$, for the test set ($newdata$) classification.
\end{enumerate}

If the user does not wish to use this phase in the classification process, the cut-off value $v$ will be $0.5$ by default. The value of 5\% was determined through preliminary tests which demonstrated that allowing a greater reduction in accuracy could misclassify a significant number of data points into a specific class.

This strategy can be employed with any fairness metric documented within the package. 

It is crucial to remember that the post-processing phase only affects a single sensitive feature. Therefore, if multiple sensitive features are utilized during the in-processing phase, just one can be selected in the post-processing phase.

The post-processing phase can be used in regular and mixed effects algorithms. In the following section, we demonstrate the effectiveness of the proposed package using multiple numerical simulations. The inputs and outputs of all post-processing functions are documented on the \href{https://github.com/JoaoVitorPamplona/FairML.jl}{package's GitHub page}.

\section{Numerical Results}
\label{sec:numerical-results}

Here, we present several numerical results to validate the proposed method's efficacy.
First, we present the step-by-step strategy used to create the synthetic datasets and to conduct the numerical experiments. The tests are run in \texttt{Julia 1.9} \parencite{BezansonEdelmanKarpinskiShah17} with the packages \texttt{JuMP} \parencite{Lubin2023}, \texttt{Ipopt} \parencite{wachter2006implementation}, to solve the optimization problems, \texttt{Distributions} \parencite{distributions} and \texttt{DataFrames} \parencite{DataFrames}. 

To create the synthetic data, we define the following parameters:

\begin{itemize}
        \item \emph{Number of points}: Number of points in the dataset; 
        \item \emph{$\beta 's$}: The fixed effects;
        \item \emph{$g's$}: The random effects with distribution $N(0,3)$, if necessary;
        \item \emph{Data points}: The covariate vector associated with fixed effects with distribution $N(0,1)$;
        \item $c$: Threshold from fair constraints;
        \item \emph{seed}: Random seed used in the generation of data;
        \item \emph{Train-Test split}: Approximately 1\% of the dataset was used for the training set, and 99\% for the test set.
\end{itemize}

The classifications of the synthetic dataset, are computed using the predictions functions \eqref{PREDEQLR}, \eqref{PREDEQSVM}, \eqref{PREDEQMELR} and \eqref{PREDEQMESVM}, depending on the problem being solved.
The package also provides these synthetic dataset generation functions. 

The tests were conducted on a laptop with an Intel Core i9-13900HX processor with a clock speed of 5.40 GHz, 64 GB of RAM, and Windows 11 operating system, with 64-bit architecture.

All figures were created using the \texttt{Plots}  and \texttt{PlotlyJS} packages, developed by \textcite{Plots} and all unspecified hyperparameters were obtained through cross-validation \parencite{browne2000cross}.

\subsection{Regular Models}
The parameters for creating synthetic datasets are as follows:
\begin{itemize}
    \item $\beta$'s $= [-2.0; 0.4; 0.8; 0.5; 2.0]$
    \item $c = 0.1$.
\end{itemize}
The $\beta_0$ is the intercept, and $\beta_4$ is the coefficient associated to the binary sensitive feature. In the unfair case, the coefficient was randomly selected using numbers between $0$ and $1$, except for $\beta_0$ and $\beta_4$. The reason for this is that we assign a high value to $\beta_4$, to give more importance to the sensitive variable in the label. In other words, data points with the sensitive categories equal to $1$ are more likely to be classified as positive. This practice results in a dataset that is inherently unfair in terms of both disparate impact and disparate mistreatment, as needed to test our methods. For all experiments, the matrix $X$ was randomly generated from a multivariate normal distribution with zero mean and independent variables. Using the generated coefficients, we employed Prediction function \eqref{PREDEQLR} to obtain labels for logistic regression tests and the prediction function in  \eqref{PREDEQSVM} for SVM tests.

In the numerical tests for regular models we consider these options of methods, all documented in Sections \ref{sec:chapter-2}, \ref{sec:chapter-3} and \ref{sec:chapter-4}:
\begin{itemize}
    \item Three options of preprocessing methods: Identity, disparate impact with $R=1$ and disparate impact with $R=5$; 
    \item Ten options of in-processing methods, logistic regression and SVM based ones;
    \item Three post-processing methods: Disparate Impact, Disparate Mistreatment and no post-processing. 
\end{itemize}
This leads to a total of $90$ scenarios with $100$ simulation runs each. For each optimization problem we impose a time limit of $60$ seconds in the in-processing stage. Only the most relevant results are shown here, the other ones can be found on \href{https://github.com/JoaoVitorPamplona/FairML.jl}{GitHub}. For each numerical test, box plots were generated for $7$ metrics.
We now present the most noteworthy numerical results. Firstly, we will demonstrate the effectiveness of the preprocessing method proposed in this work.

\begin{figure}[H]
\centering
\begin{subfigure}{0.49\textwidth}
\includegraphics[scale=.33]{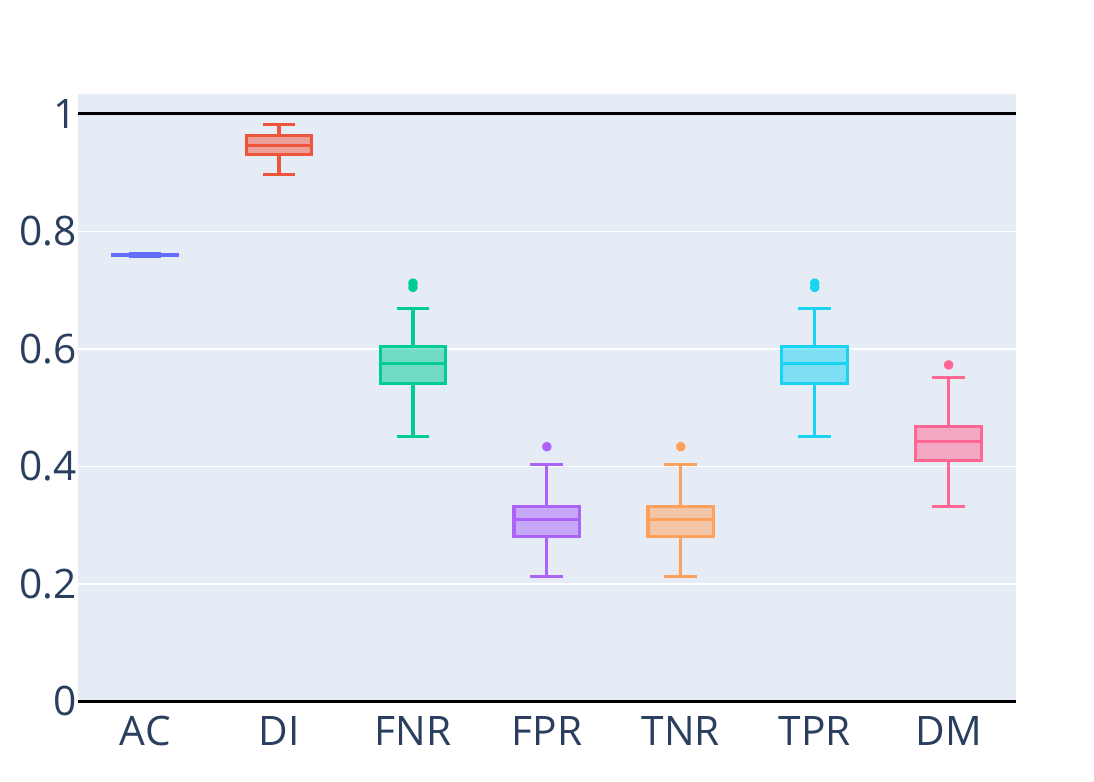}
\end{subfigure}
\begin{subfigure}{0.49\textwidth}
\includegraphics[scale=.33]{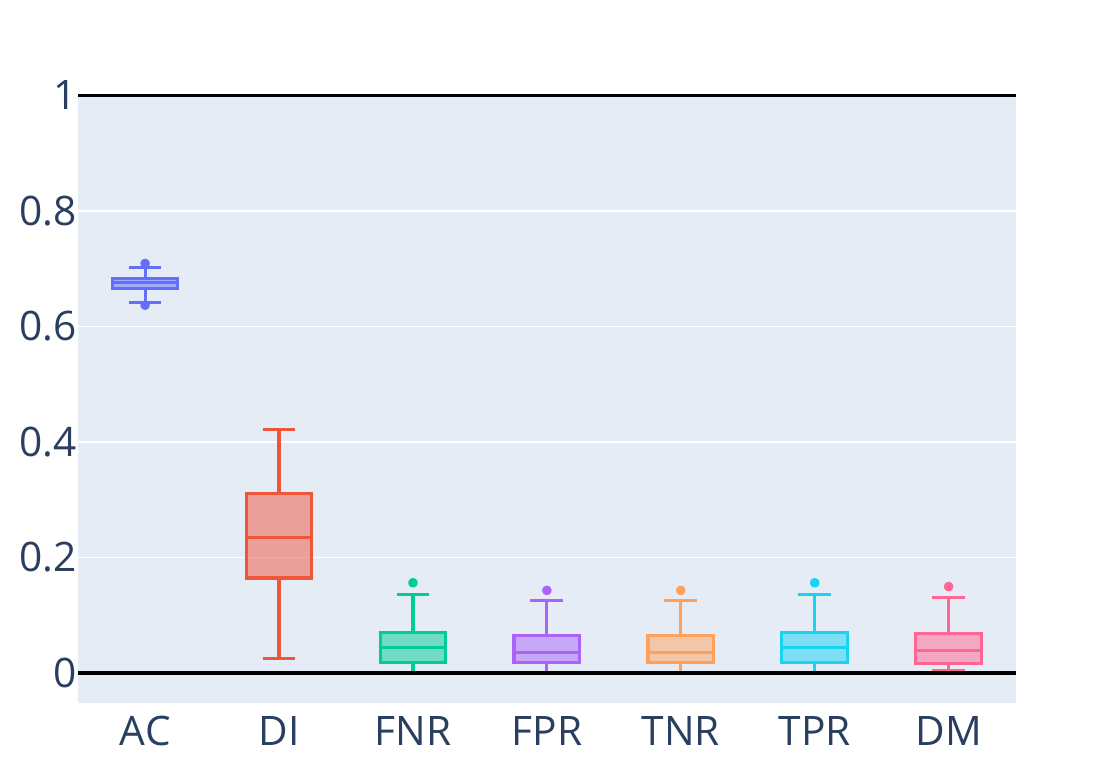}
\end{subfigure}
\end{figure}
\begin{figure}[H]\ContinuedFloat
\begin{subfigure}{0.49\textwidth}
\includegraphics[scale=.33]{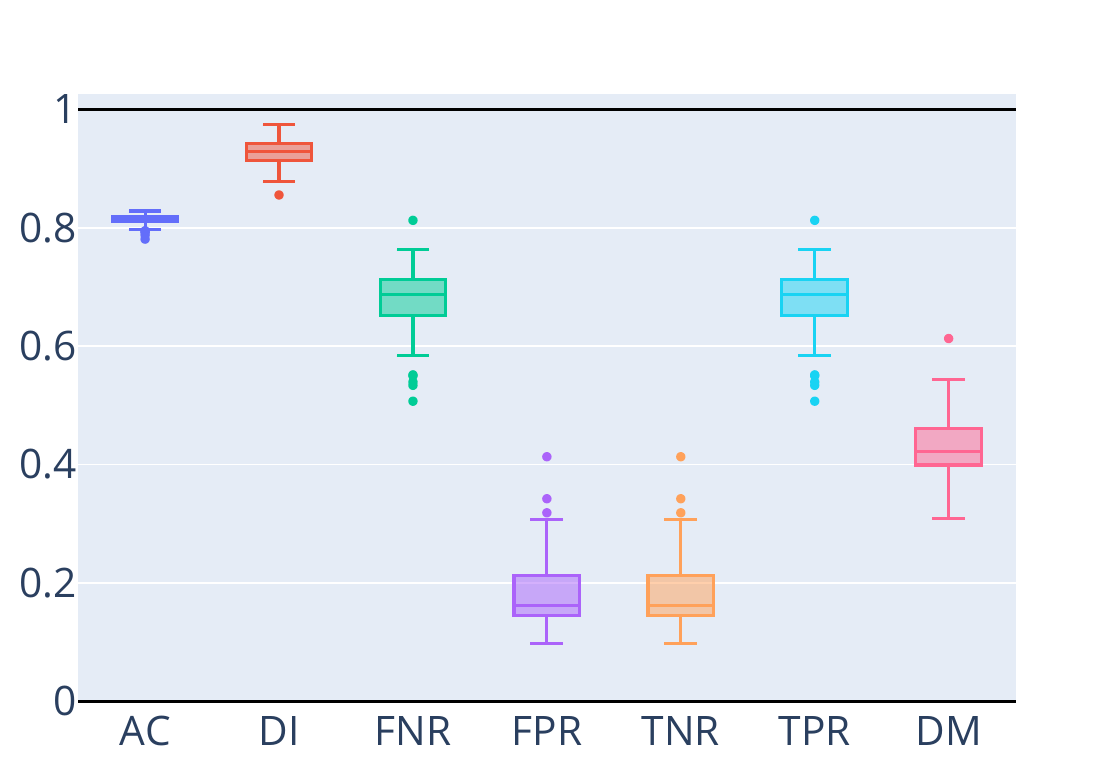}
\end{subfigure}
\begin{subfigure}{0.49\textwidth}
\includegraphics[scale=.33]{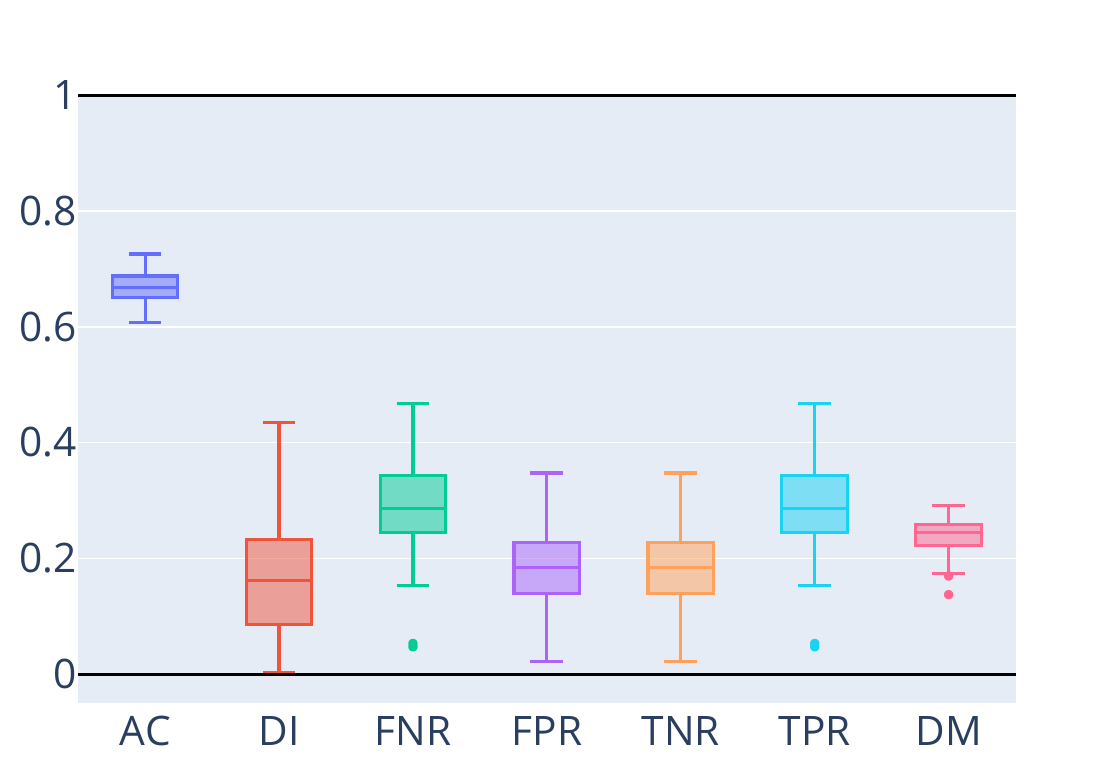}
\end{subfigure}
\caption{Preprocessing results: First row: Comparison for logistic regression. Second row: Comparison for support vector machine. Left: Without preprocessing. Right: With preprocessing (R=1)}
\label{fig1}
\end{figure}

As can be seen in Figure \ref{fig1} for both logistic regression and SVM, the proposed resampling method, significantly reduces the disparate impact. It is also worth noting that this leads to a decline in other fairness metrics as well. This implies in a decrease of accuracy, however, this is an anticipated outcome in the field of fair machine learning. Now, considering the same preprocessing but being executed multiple times:
\begin{figure}[H]
\centering
\begin{subfigure}{0.49\textwidth}
\includegraphics[scale=.33]{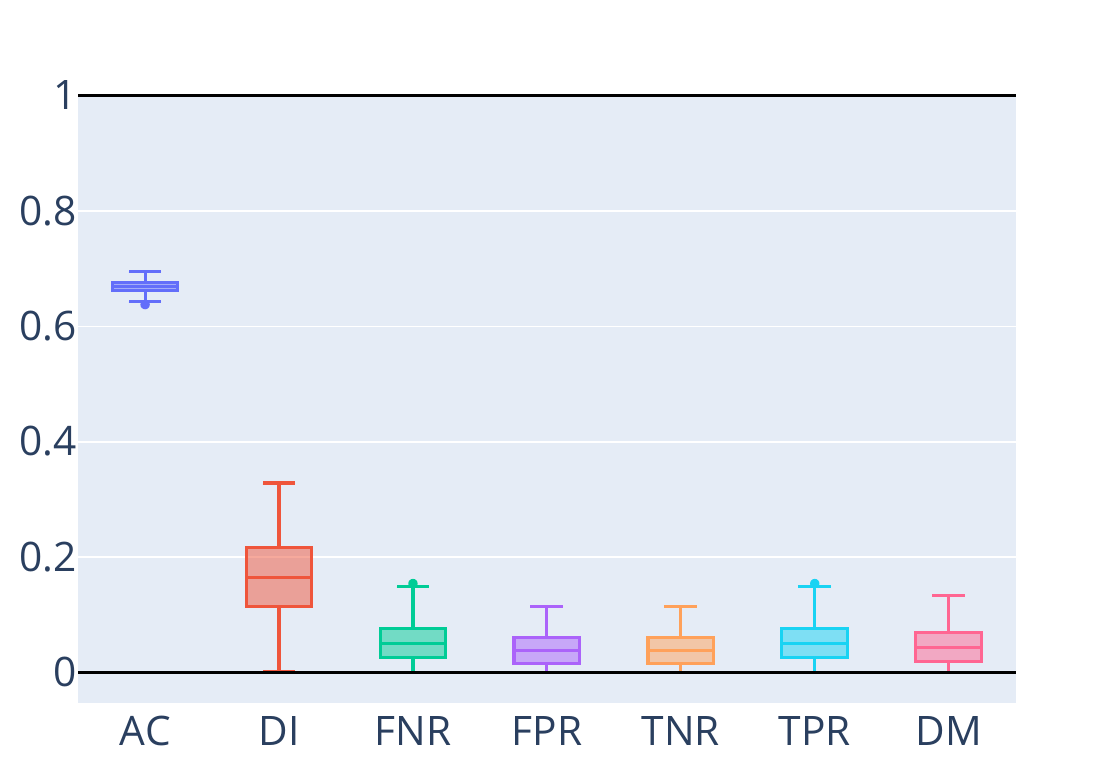}
\end{subfigure}
\begin{subfigure}{0.49\textwidth}
\includegraphics[scale=.33]{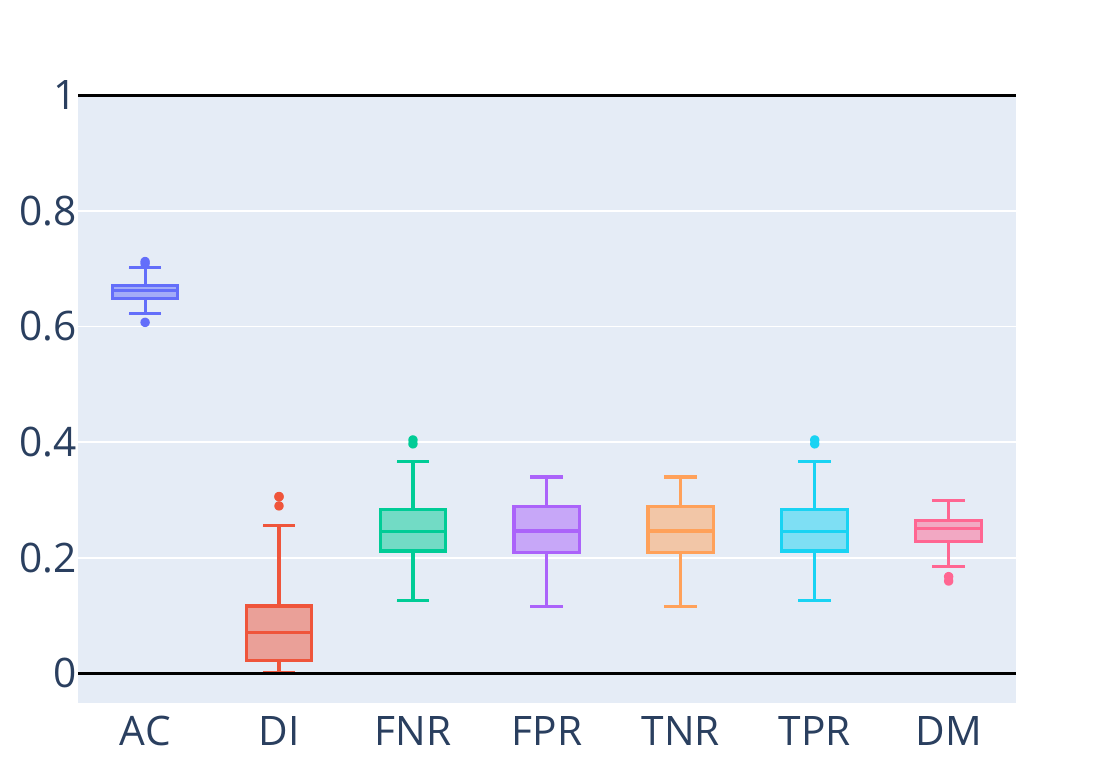}
\end{subfigure}
\caption{Preprocessing with multiple runs (R=5): Left: Logistic regression. Right: Support vector machine.}
\label{fig5}
\end{figure}
It can be observed from Figure \ref{fig5} that repeating the resampling method and selecting the best solution is also an effective approach, in comparison to the right side of the Figure \ref{fig1}, which is executed only once. Therefore, it is recommended when time is not an issue.

Henceforth, the following numerical tests focus on optimization problems during the in-processing phase.

\begin{figure}[H]
\centering
\begin{subfigure}{0.49\textwidth}
\includegraphics[scale=.33]{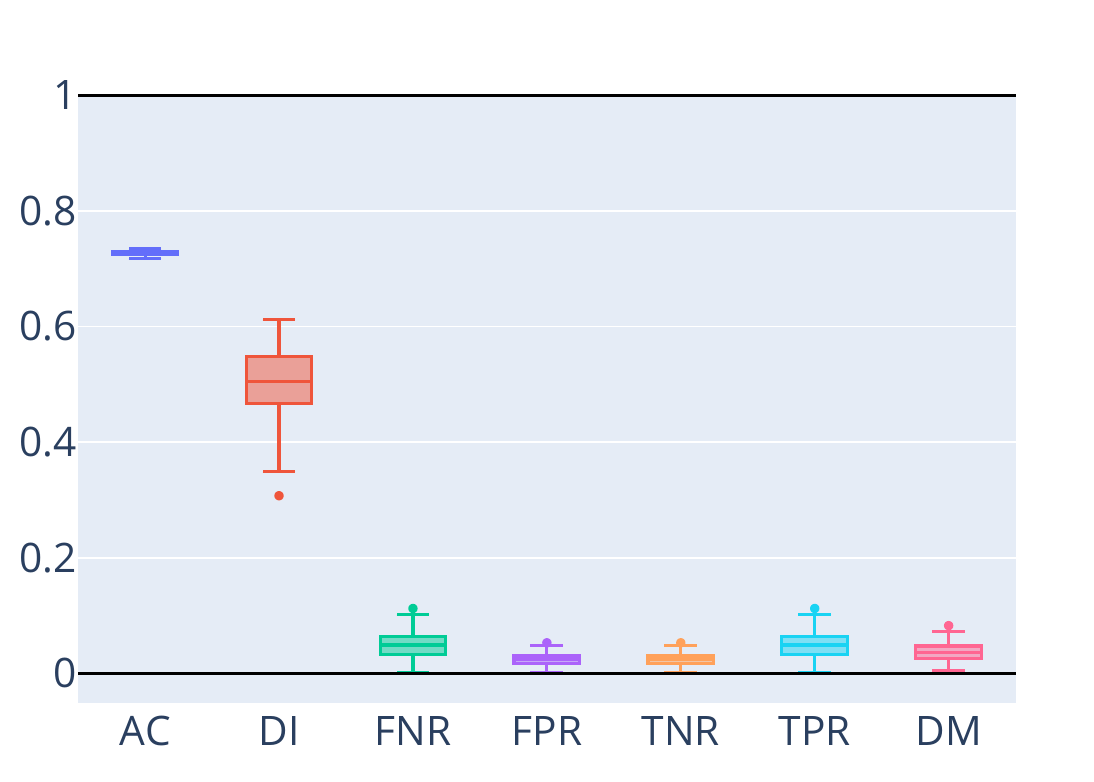}
\end{subfigure}
\begin{subfigure}{0.49\textwidth}
\includegraphics[scale=.33]{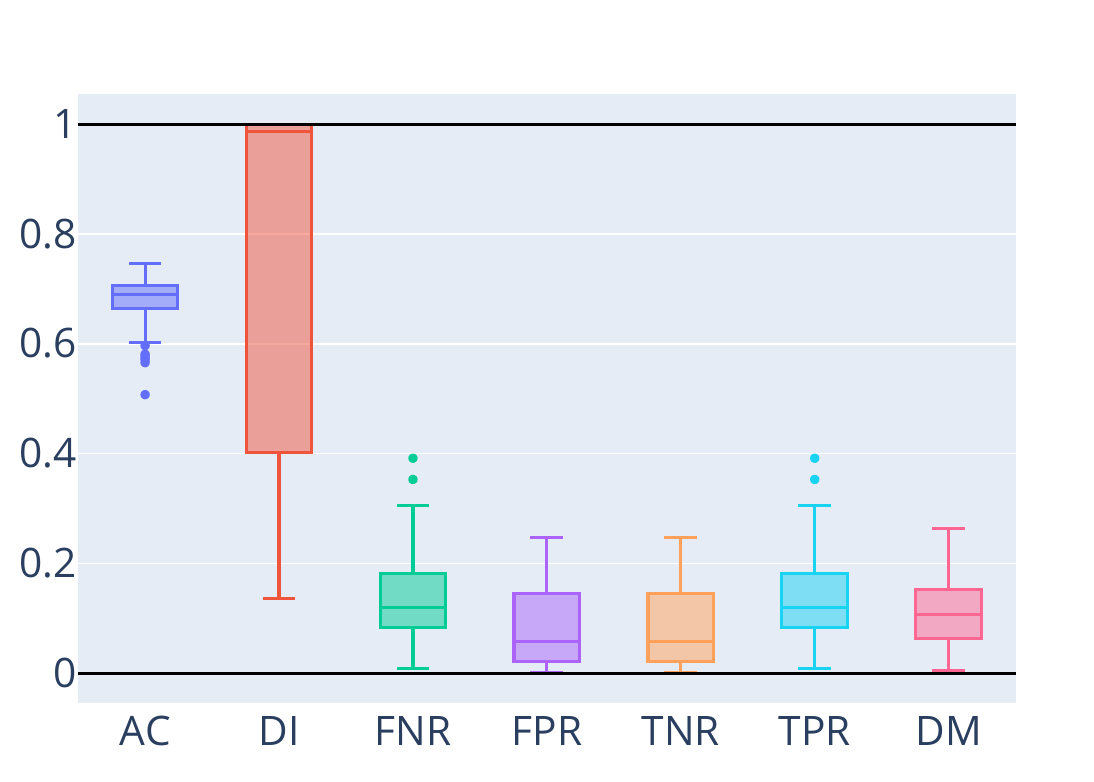}
\end{subfigure}
\end{figure}
\begin{figure}[H]\ContinuedFloat
\begin{subfigure}{0.49\textwidth}
\includegraphics[scale=.33]{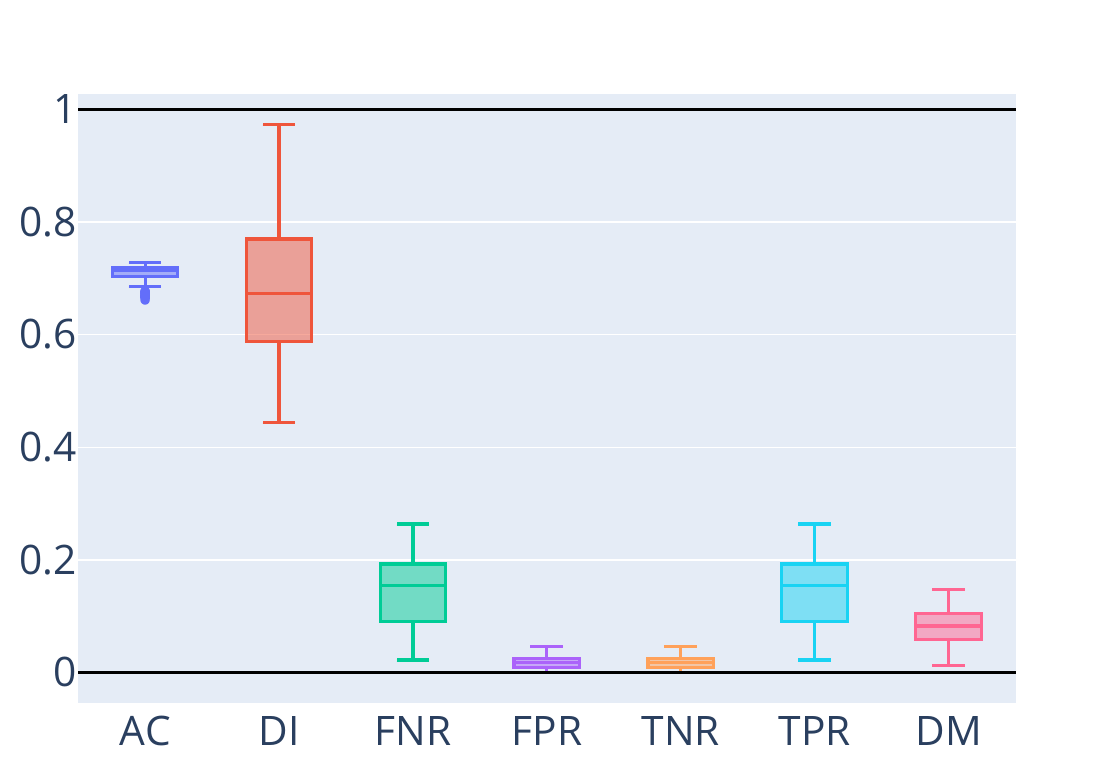}
\end{subfigure}
\begin{subfigure}{0.49\textwidth}
\includegraphics[scale=.33]{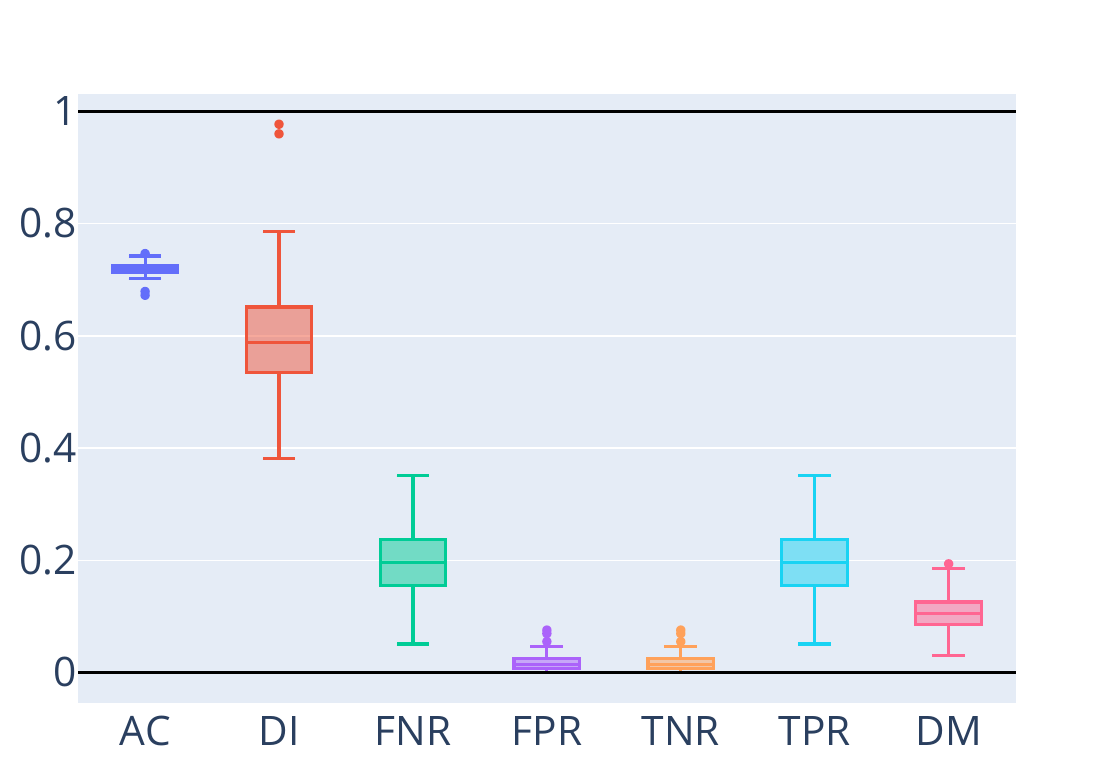}
\end{subfigure}
\caption{In-processing results: First row: Comparison for logistic regression. Second row: Comparison for support vector machine. Left: Disparate impact. Right: Disparate mistreatment.}
\label{fig8}
\end{figure}

In this set of tests, we can verify that, when compared to tests without fairness constraints, in Figures \ref{fig1}, the fair optimization problems effectively reduced the fairness metrics they are designed to mitigate. I.e., when using the optimization problems with disparate impact constraints we have a decrease of DI. We can see similar results for disparate mistreatment.

Finally, we demonstrate the effectiveness of the post-processing phase, also proposed in this paper.

\begin{figure}[H]
\centering
\begin{subfigure}{0.49\textwidth}
\includegraphics[scale=.33]{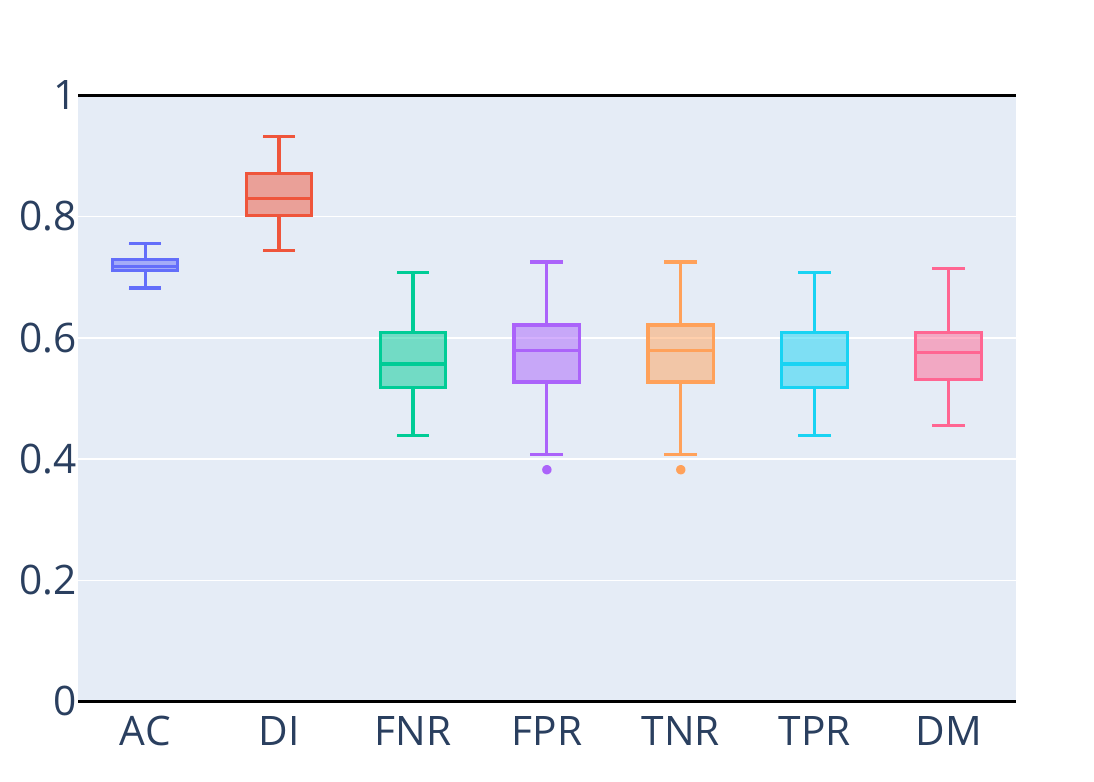}
\end{subfigure}
\begin{subfigure}{0.49\textwidth}
\includegraphics[scale=.33]{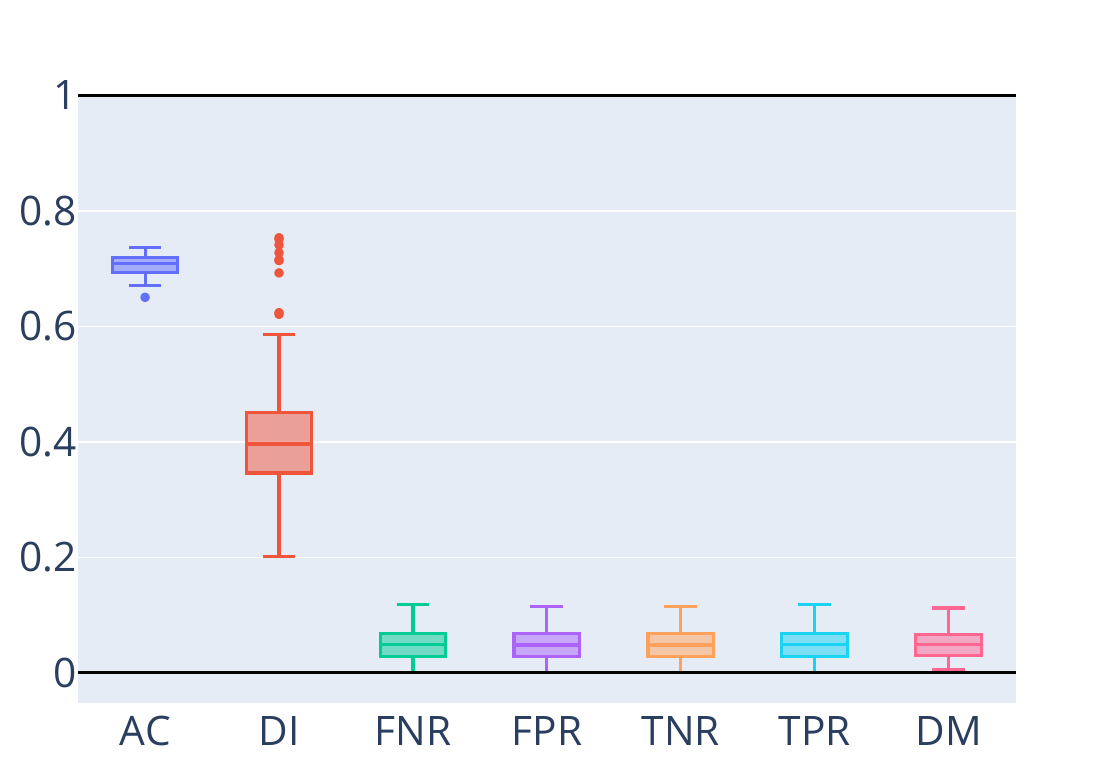}
\end{subfigure}
\end{figure}
\begin{figure}[H]\ContinuedFloat
\begin{subfigure}{0.49\textwidth}
\includegraphics[scale=.33]{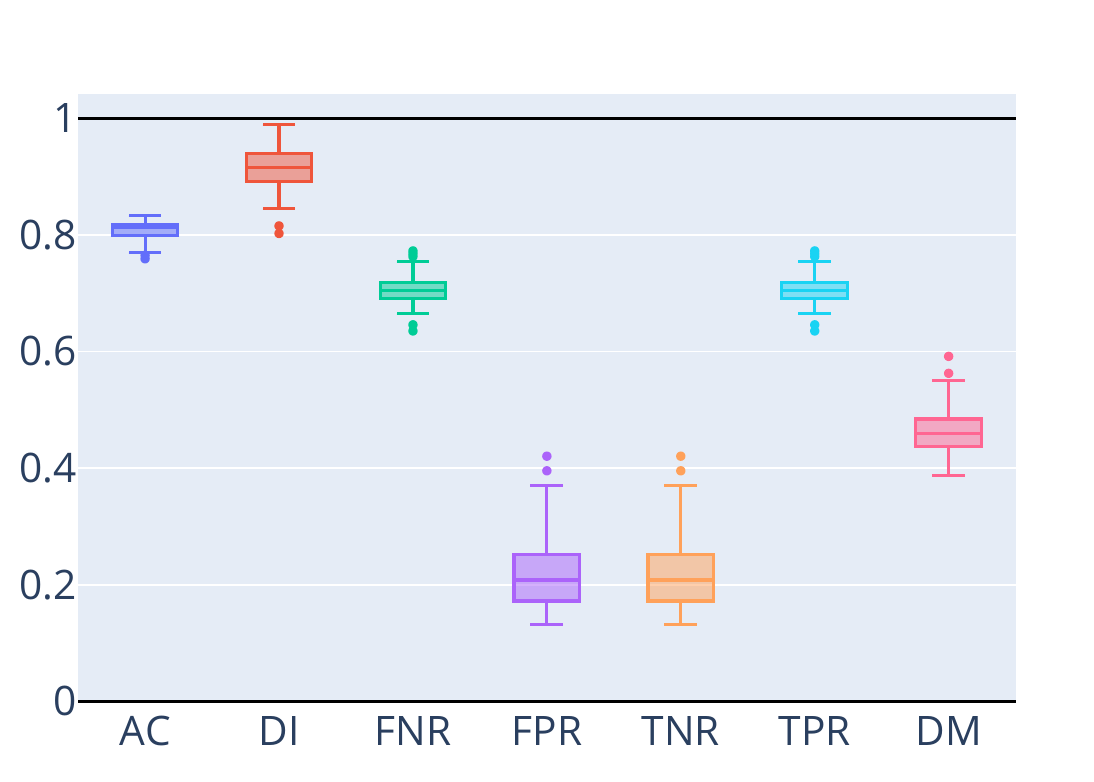}
\end{subfigure}
\begin{subfigure}{0.49\textwidth}
\includegraphics[scale=.33]{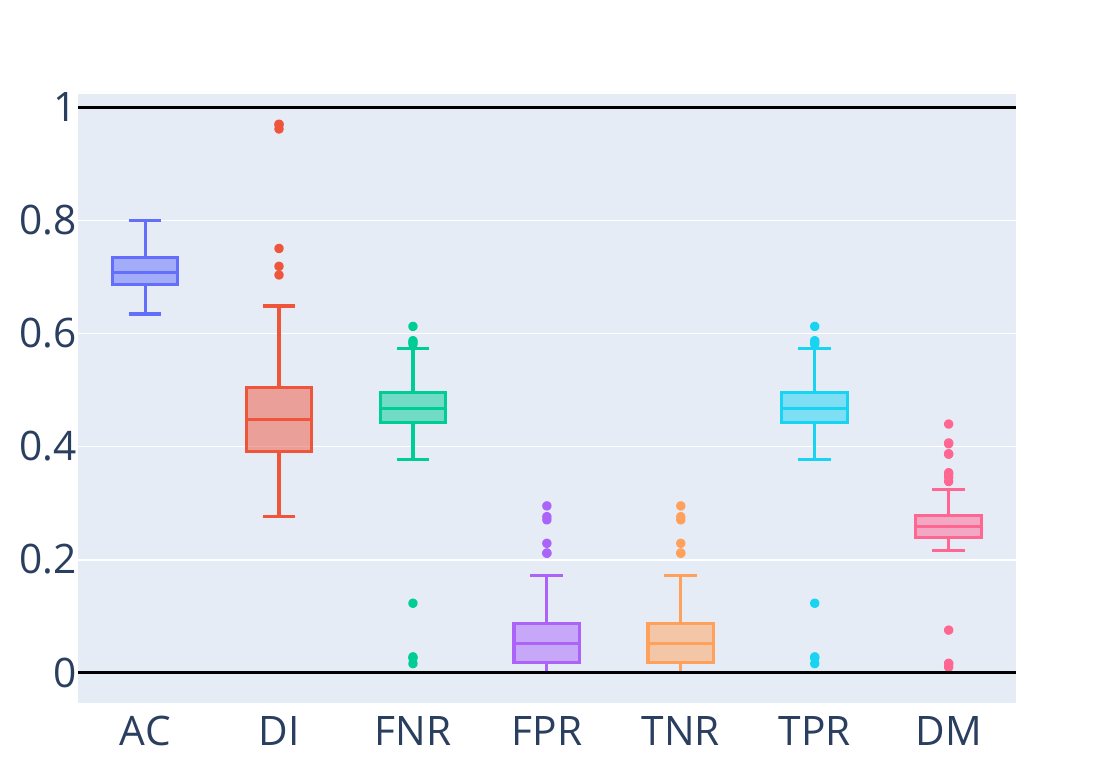}
\end{subfigure}
\caption{Post-processing results for disparate impact: First row: Comparison for logistic regression. Second row: Comparison for support vector machine. Left: Only post-processing. Right: In-processing and post-processing.}
\label{fig11}
\end{figure}

\begin{figure}[H]
\centering
\begin{subfigure}{0.49\textwidth}
\includegraphics[scale=.33]{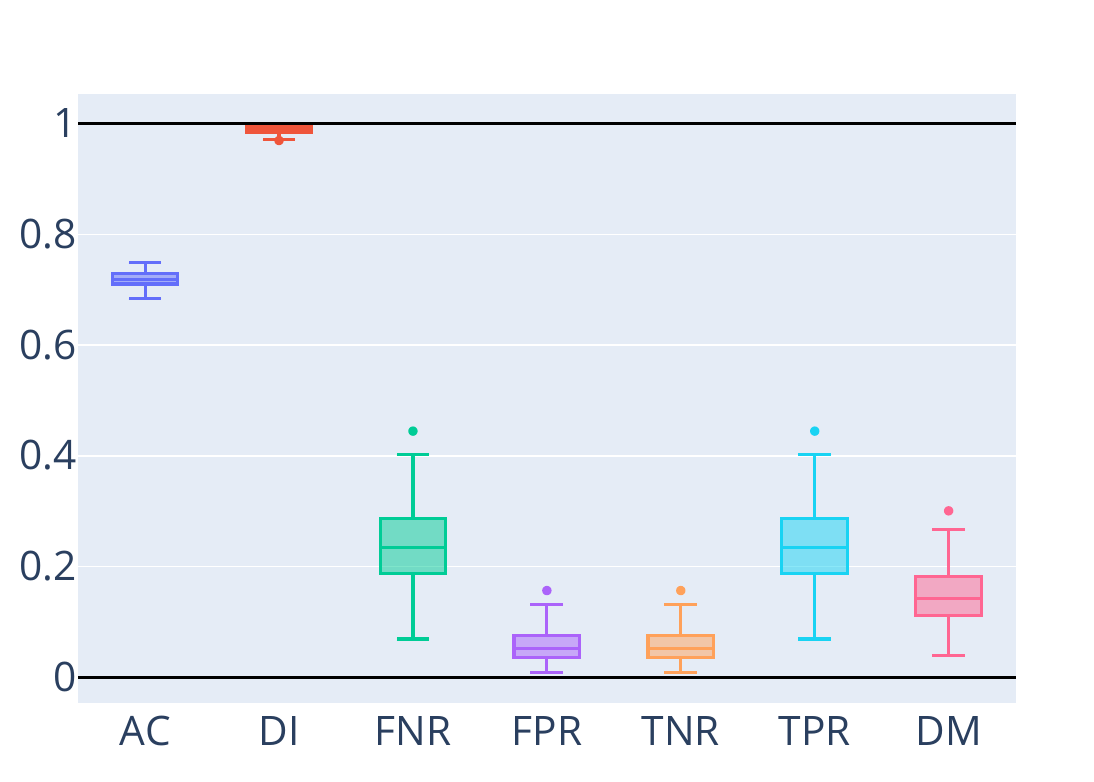}
\end{subfigure}
\begin{subfigure}{0.49\textwidth}
\includegraphics[scale=.33]{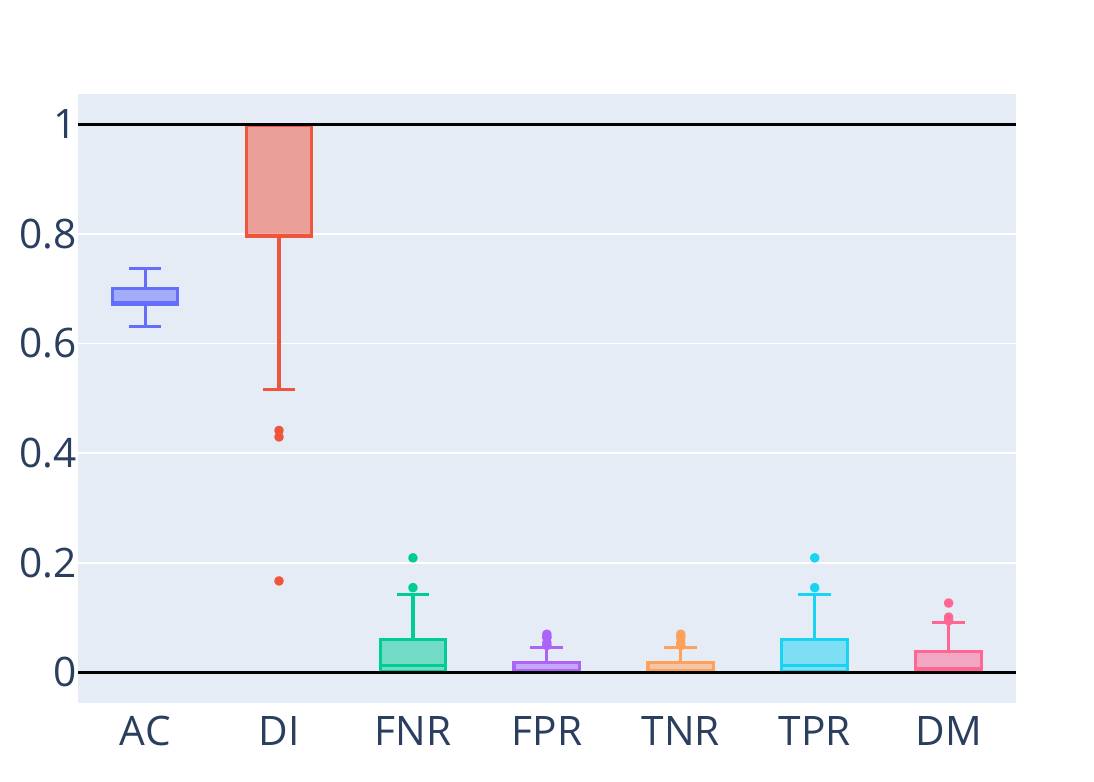}
\end{subfigure}
\end{figure}
\begin{figure}[H]\ContinuedFloat
\begin{subfigure}{0.49\textwidth}
\includegraphics[scale=.33]{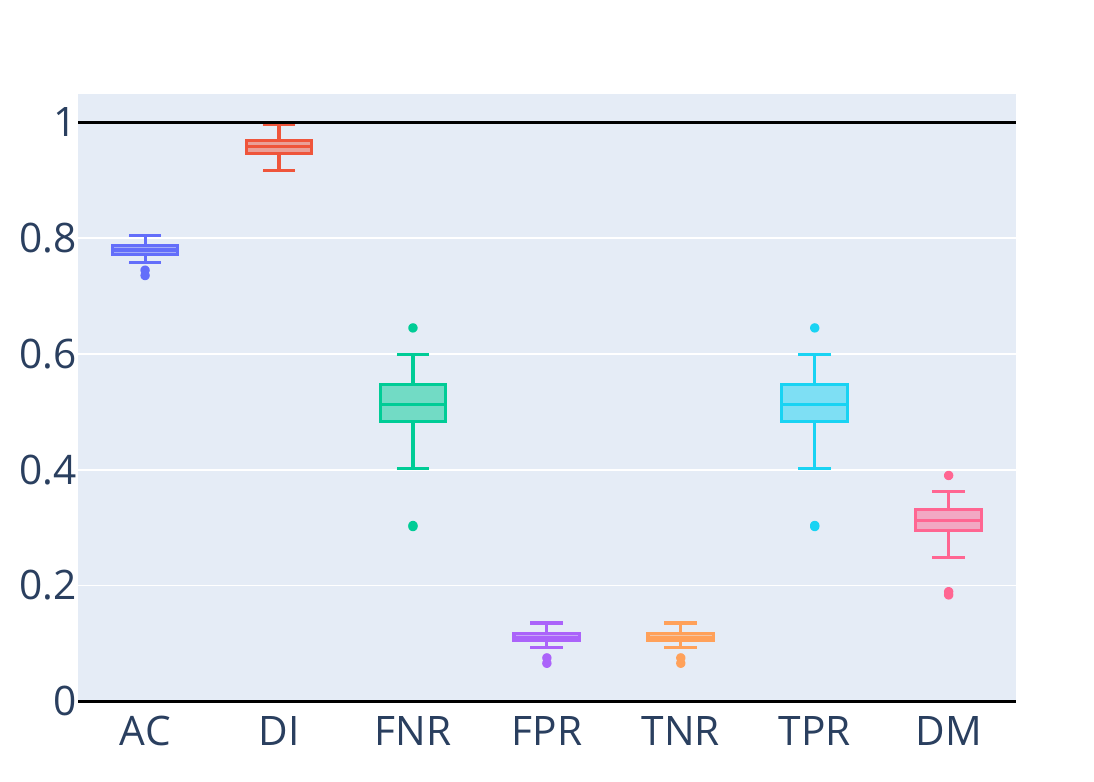}
\end{subfigure}
\begin{subfigure}{0.49\textwidth}
\includegraphics[scale=.33]{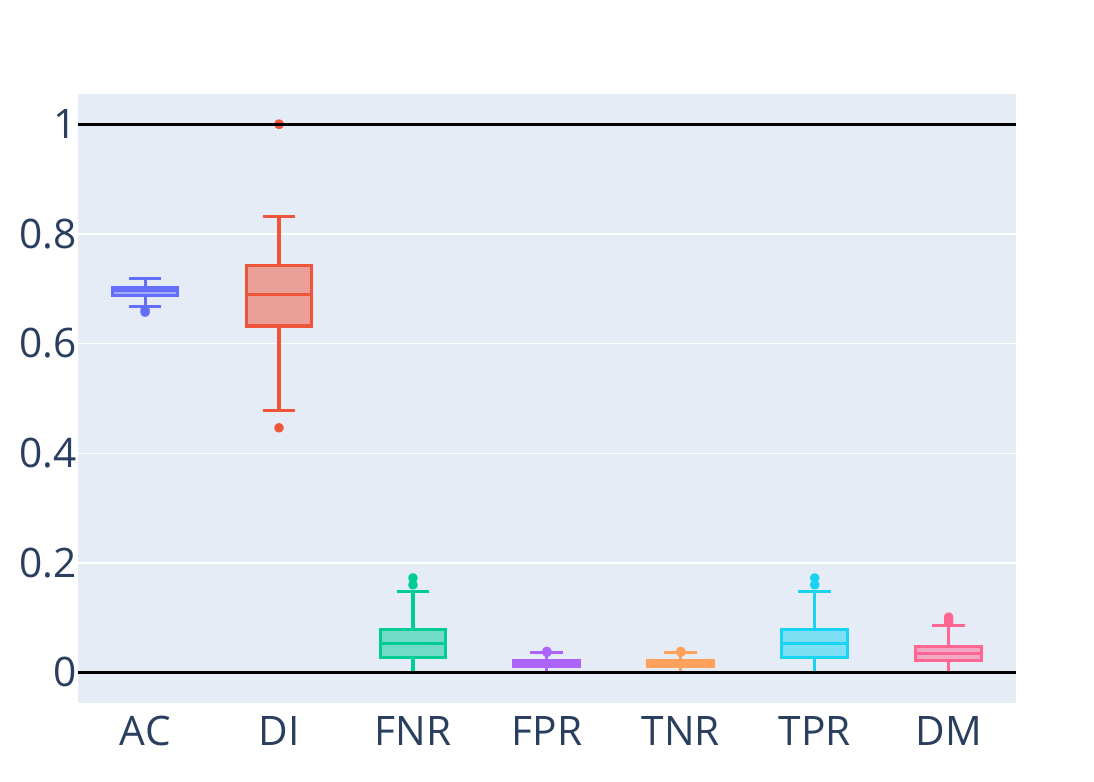}
\end{subfigure}
\caption{Post-processing results for disparate mistreatment: First row: Comparison for logistic regression. Second row: Comparison for support vector machine. Left: Only post-processing. Right: In-processing and post-processing.}
\label{fig13}
\end{figure}

The post-processing phase can be utilized independently, without the in-processing phase affecting the fair metrics, or both phases can be employed simultaneously.
It can be observed, in the left side of Figures \ref{fig11} and \ref{fig13}, that employing solely the post-processing phase leads to an slight improvement in the desired fairness metric without significantly compromising accuracy.
Furthermore, it can be noted that utilizing both strategies in conjunction, as can be seen in the right side of Figures \ref{fig11} and \ref{fig13}, yields superior outcomes compared to employing either in-processing or post-processing alone. Consequently, our recommendation is to utilize both strategies simultaneously.

\subsection{Mixed Models}
The parameters for creating synthetic datasets are as follows:
\begin{itemize}
    \item $\beta$'s $= [-4.0;0.4;0.8;0.5;4.0]$;
    \item $g$'s: 100 groups with $b_i \sim N(0,3.0)$, with $i \in [1, 100]$;
    \item $c = 0.1$.
\end{itemize}
In the numerical tests for mixed models we consider the following options of methods, documented in Sections \ref{sec:chapter-3} and \ref{sec:chapter-4}:
\begin{itemize}
    \item Ten options of in-processing methods, logistic regression and SVM based ones;
    \item Three post-processing methods: Disparate Impact, Disparate Mistreatment and no post-processing. 
\end{itemize}
Hence, we have $30$ scenarios, with $100$ simulation runs each. For each optimization problem we impose a time limit of $60$ seconds in the in-processing stage. As for regular models, we only present the most important results, the rest can be found on \href{https://github.com/JoaoVitorPamplona/FairML.jl}{GitHub}. Since the mixed models strategy does not include a preprocessing phase, we will first examine the in-processing phase, where the optimization problems proposed in this work are solved.

\begin{figure}[H]
\centering
\begin{subfigure}{0.49\textwidth}
\includegraphics[scale=.33]{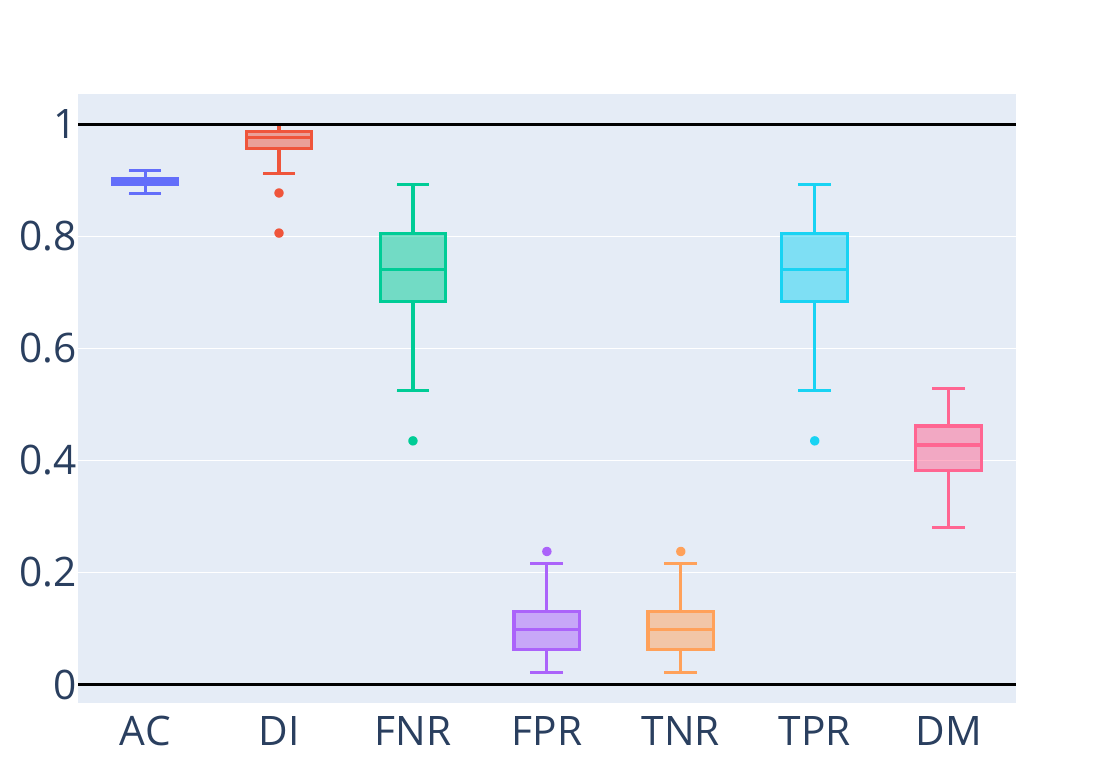}
\end{subfigure}
\begin{subfigure}{0.49\textwidth}
\includegraphics[scale=.33]{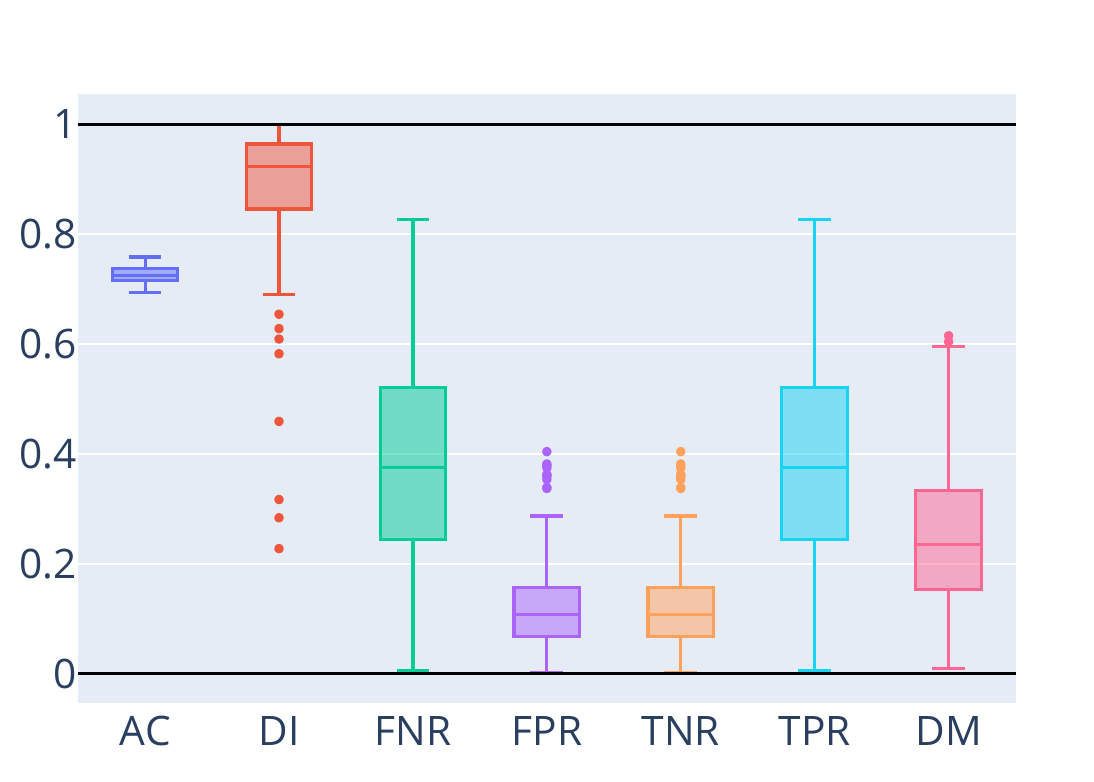}
\end{subfigure}
\begin{subfigure}{0.49\textwidth}
\includegraphics[scale=.33]{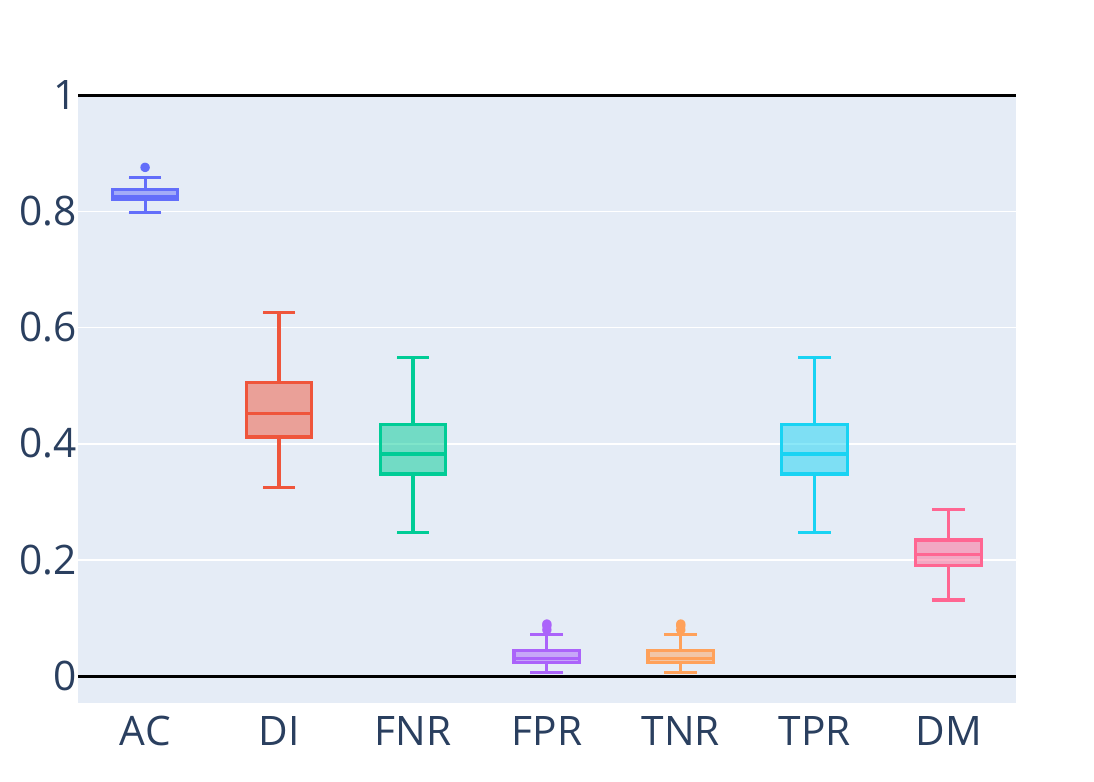}
\end{subfigure}
\begin{subfigure}{0.49\textwidth}
\includegraphics[scale=.33]{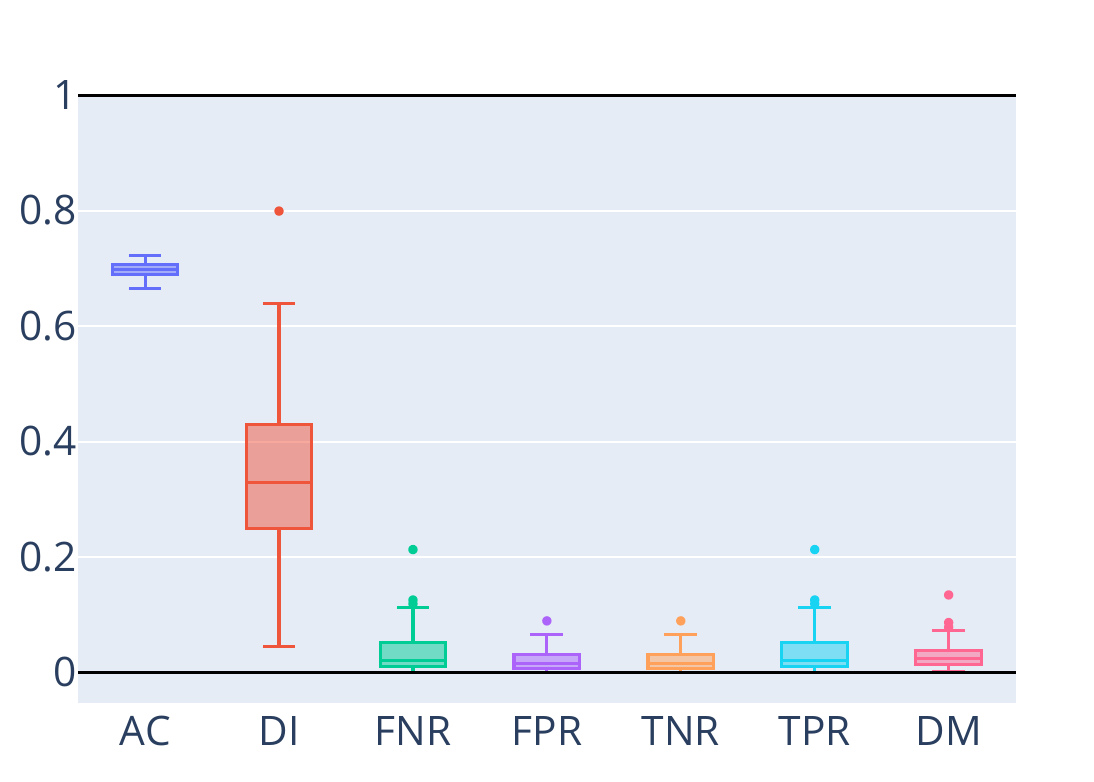}
\end{subfigure}
\end{figure}
\begin{figure}[H]\ContinuedFloat
\begin{subfigure}{0.49\textwidth}
\includegraphics[scale=.33]{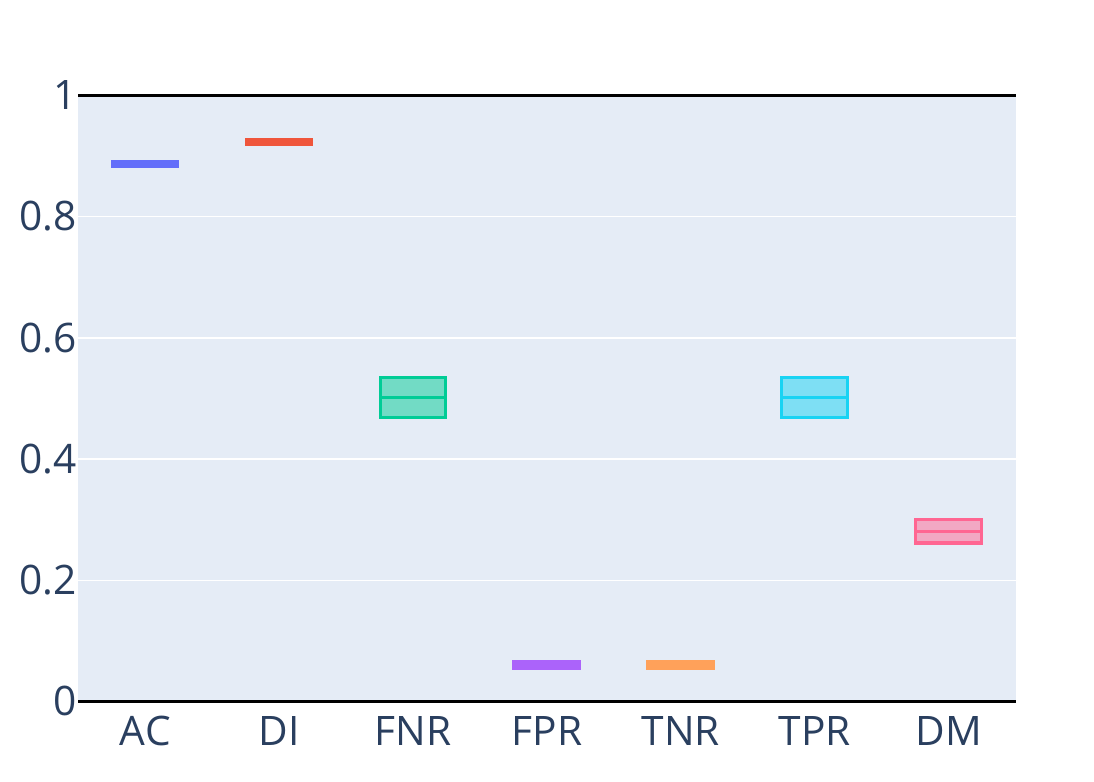}
\end{subfigure}
\begin{subfigure}{0.49\textwidth}
\includegraphics[scale=.33]{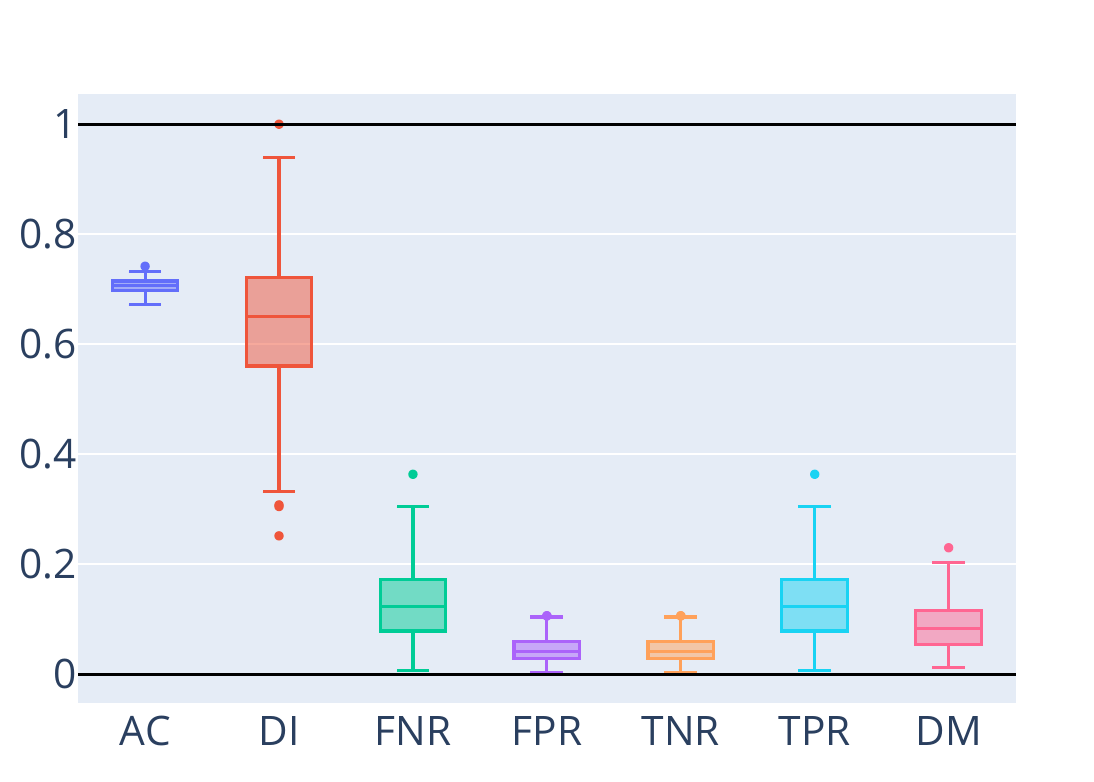}
\end{subfigure}
\caption{In-processing results in Mixed models: Left: Comparison for logistic regression. Right: Comparison for support vector machine. First row: No fairness constraints. Second row: Disparate impact constraints. Third row: Disparate mistreatment constraints.}
\label{fig19}
\end{figure}

Figure \ref{fig19} confirms that fairness constraints in the optimization problems successfully improve the fairness metrics they were designed to address. That is, when incorporating disparate impact constraints into optimization problems, we observe a reduction in disparate impact. Similar results are evident for disparate mistreatment.

At last, we show the effectiveness of the post-processing step on the mixed models.

\begin{figure}[H]
\centering
\begin{subfigure}{0.49\textwidth}
\includegraphics[scale=.33]{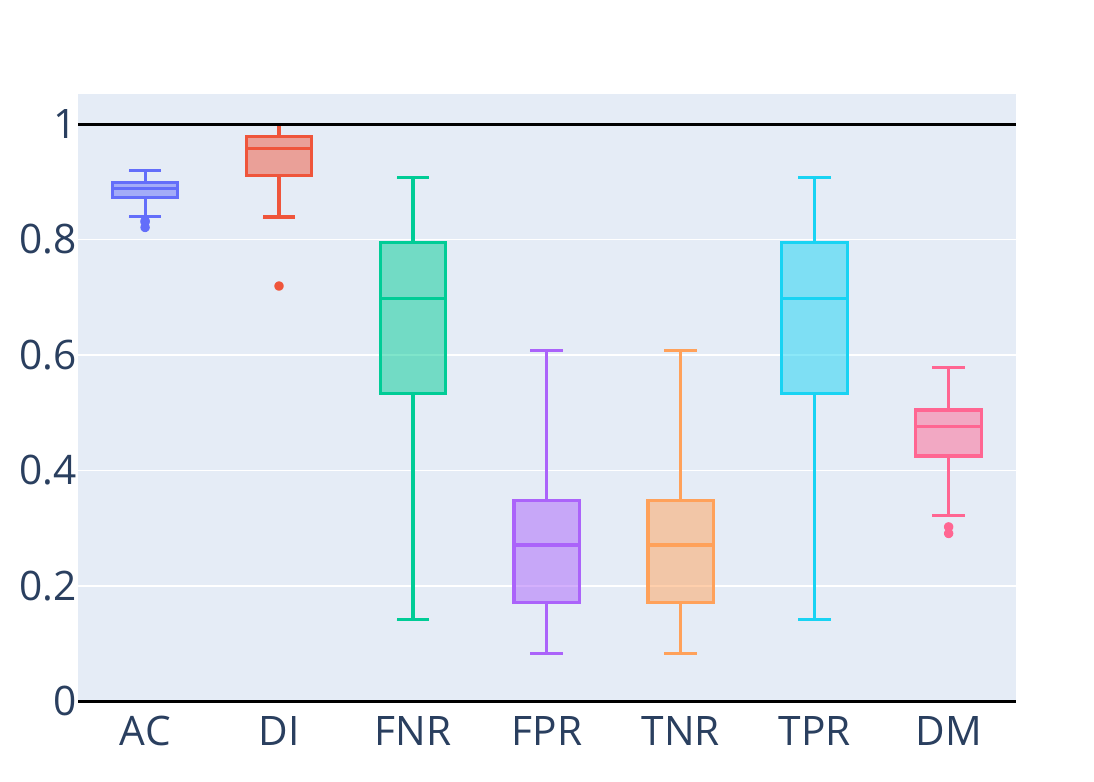}
\end{subfigure}
\begin{subfigure}{0.49\textwidth}
\includegraphics[scale=.33]{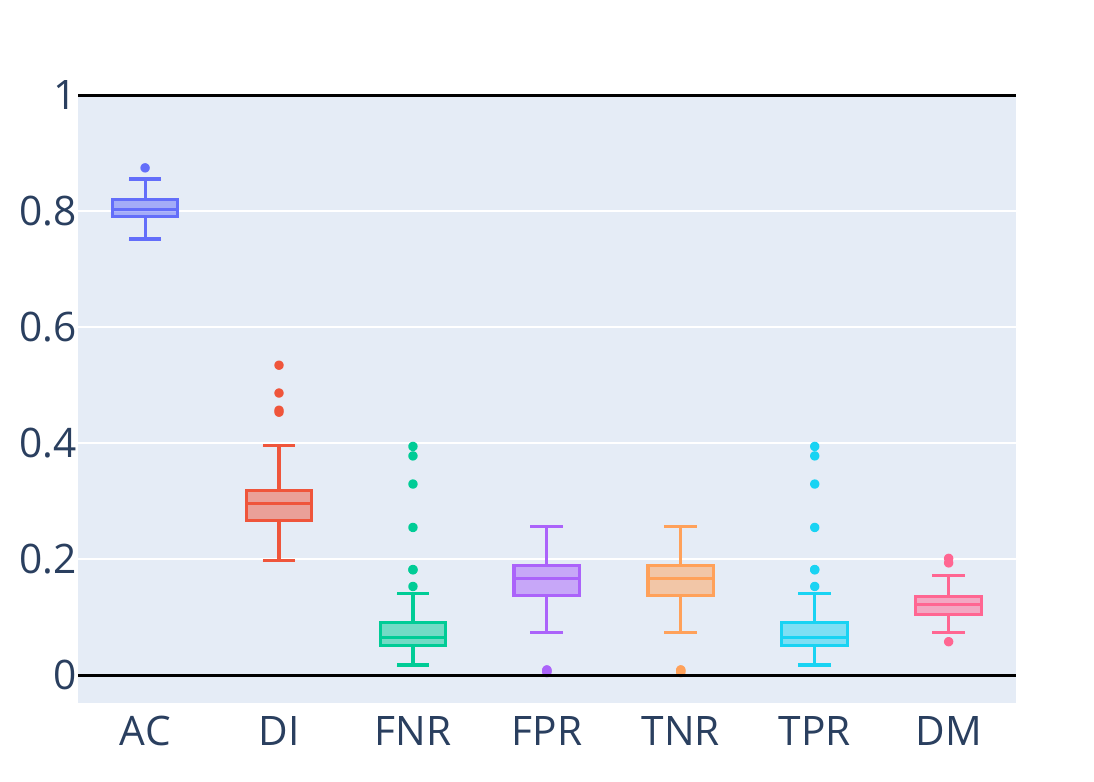}
\end{subfigure}
\begin{subfigure}{0.49\textwidth}
\includegraphics[scale=.33]{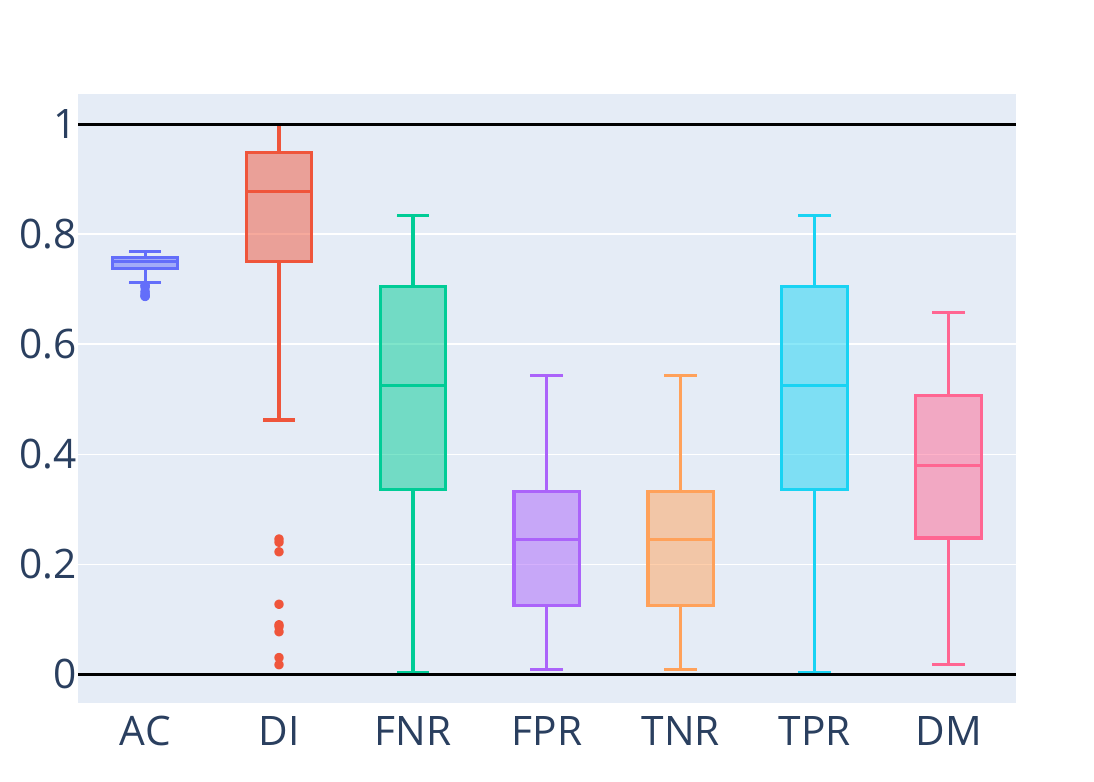}
\end{subfigure}
\begin{subfigure}{0.49\textwidth}
\includegraphics[scale=.33]{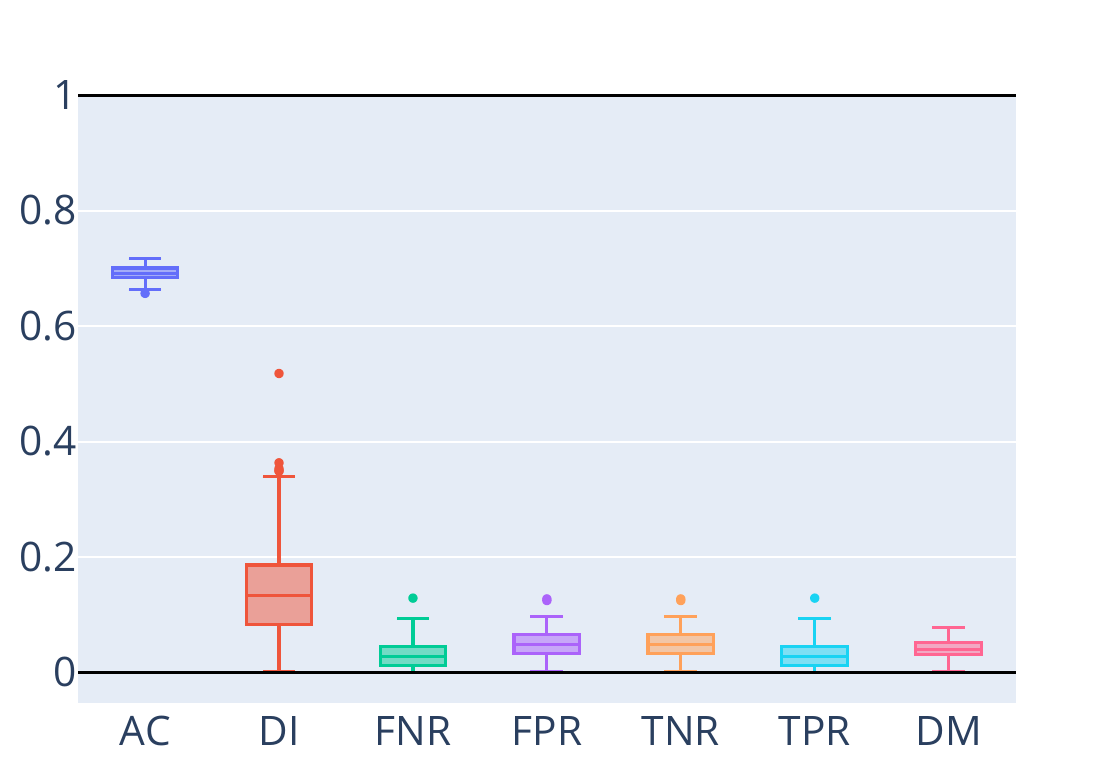}
\end{subfigure}
\caption{Post-processing results in mixed models for disparate impact: First row: Comparison for logistic regression. Second row: Comparison for support vector machine. Left: Only post-processing. Right: In-processing and post-processing.}
\label{fig25}
\end{figure}

Just as in post-processing tests for regular models, while the post-processing phase can function independently, its integration with the in-processing phase yields superior results. Consequently, we reiterate our recommendation to employ both phases simultaneously as can be seen in Figures \ref{fig25} and \ref{fig26}.

\begin{figure}[H]
\begin{subfigure}{0.49\textwidth}
\includegraphics[scale=.33]{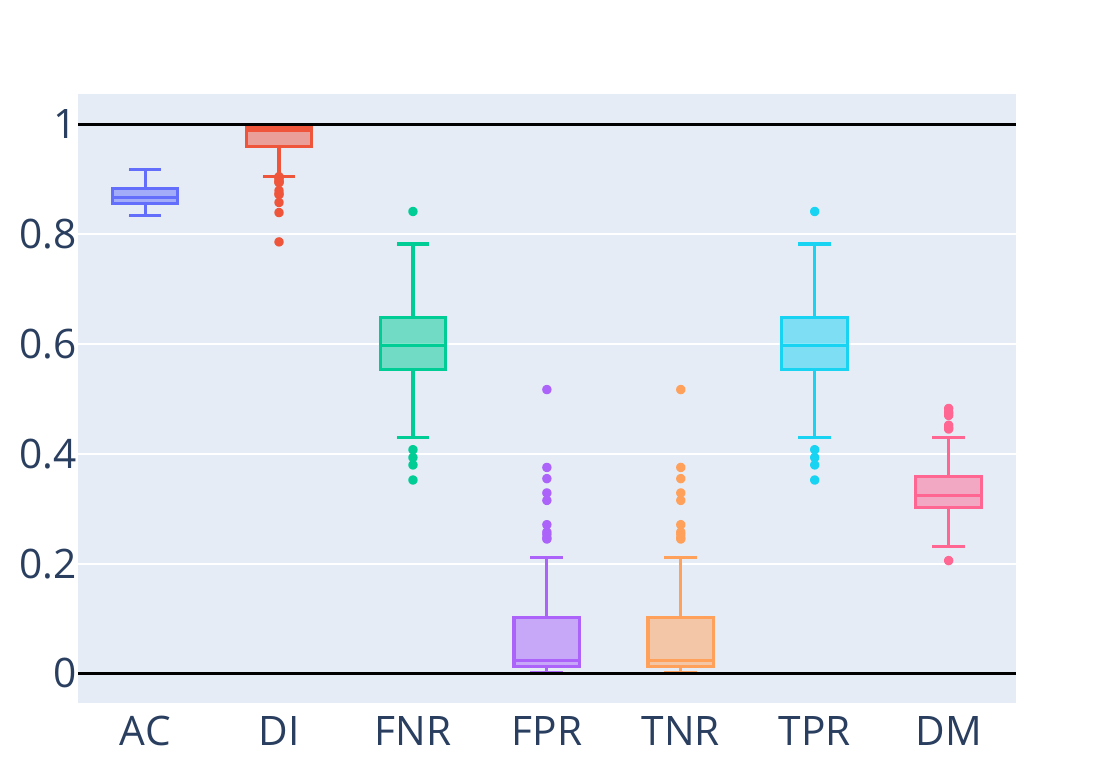}
\end{subfigure}
\begin{subfigure}{0.49\textwidth}
\includegraphics[scale=.33]{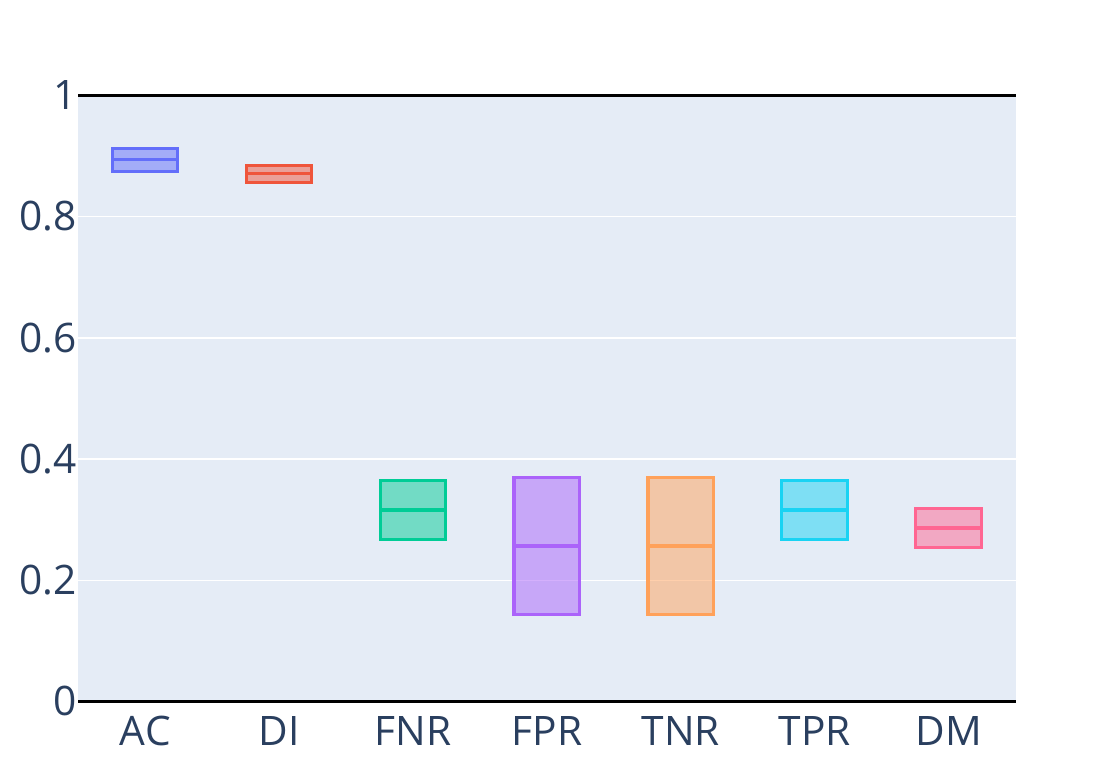}
\end{subfigure}
\end{figure}
\begin{figure}[H]\ContinuedFloat
\begin{subfigure}{0.49\textwidth}
\includegraphics[scale=.33]{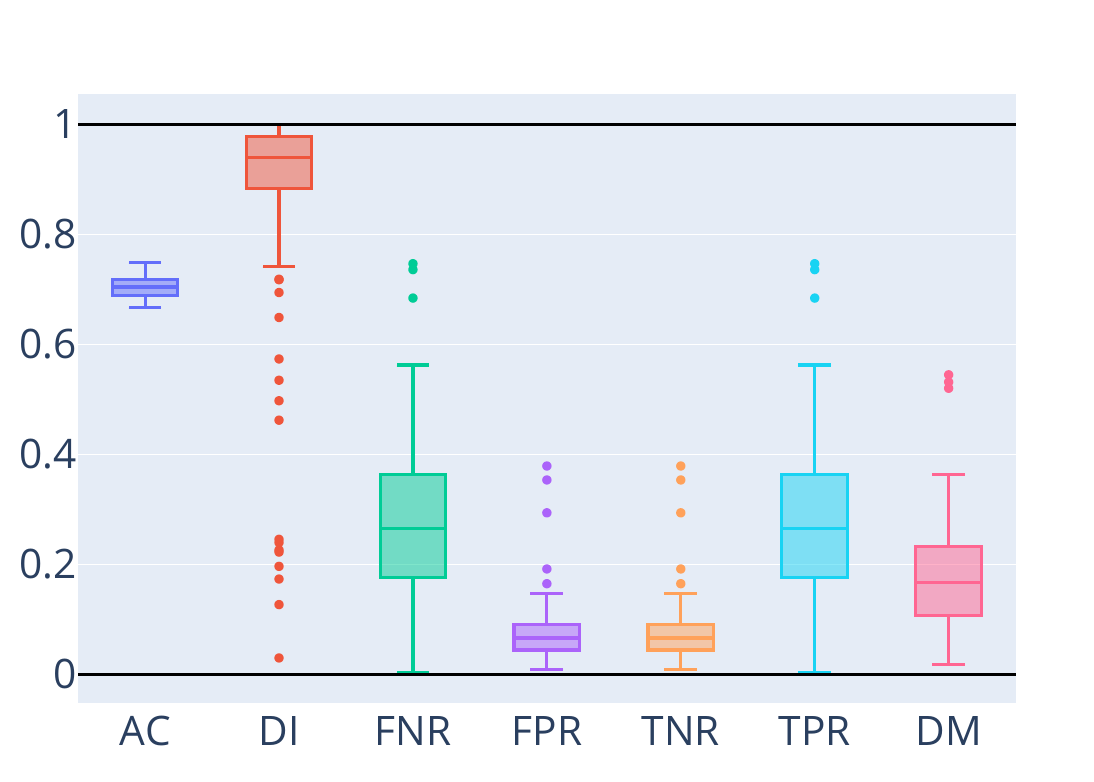}
\end{subfigure}
\begin{subfigure}{0.49\textwidth}
\includegraphics[scale=.33]{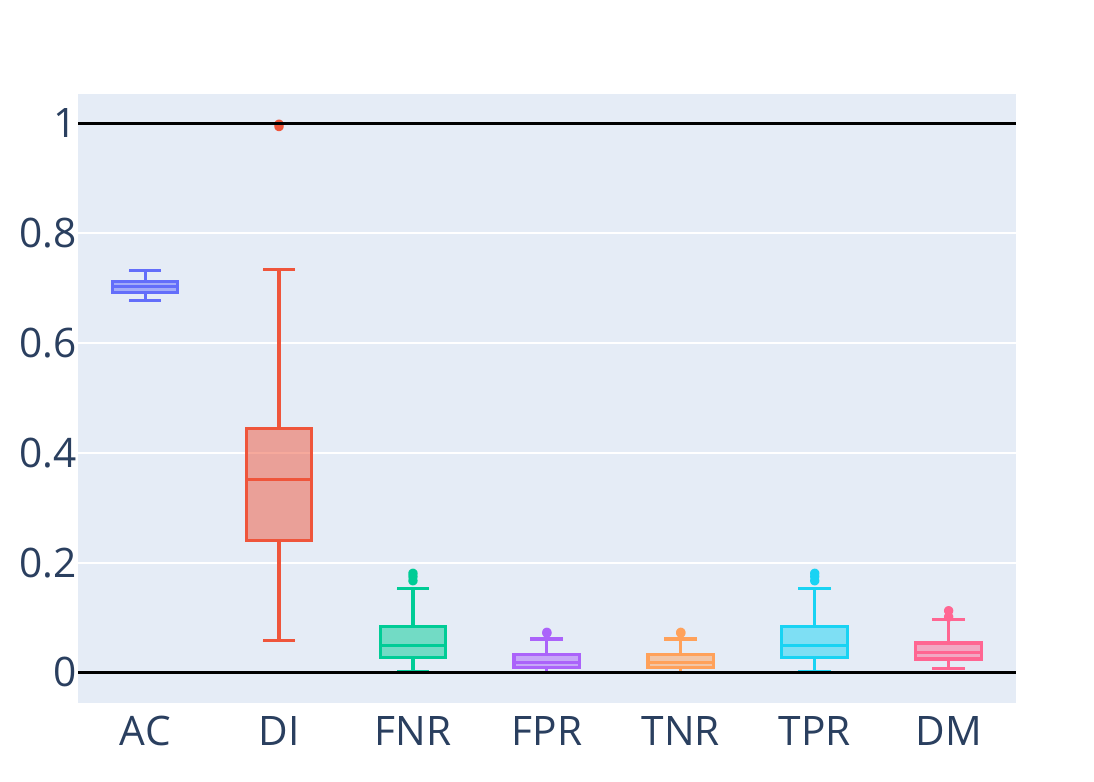}
\end{subfigure}
\caption{Post-processing results in mixed models for disparate mistreatment: First row: Comparison for logistic regression. Second row: Comparison for support vector machine. Left: Only post-processing. Right: In-processing and post-processing.}
\label{fig26}
\end{figure}

\section{Conclusion}
\label{sec:conclusion}

In this work we propose \texttt{FairML.jl} a \texttt{Julia} package that addresses fairness for classification in machine learning, offering a versatile tools to tackle unfairness at various stages of the classification process, providing users with more choices and control.

The first step, called preprocessing phase, employs a resampling method to mitigate disparate impact. This method utilizes a mixed strategy that combines undersampling and cross-validation.

In the in-processing phase, we extended the original optimization problems of support vector machine and logistic regression to address unfairness in the presence of group bias within the data. Specifically, we propose constrained optimization models that mitigate unfairness in heterogeneous populations. This phase also allows the utilization of any binary classifier from package \texttt{MLJ.jl} as learning tool.

Additionally, this paper proposes a post-processing method designed to identify a solution that improves the specified fairness metric, given by the user, without significantly compromising accuracy.

With simulations, we showcased how our approach reduces unfairness in the three phases. We also conducted some cross-phase combinations that can further enhance the final solutions.

To improve the framework's capabilities, future work focuses on incorporating additional fairness metrics, and to adapt the phases to deal with multiclass classification.
\section*{Acknowledgements}

The authors are grateful for the support of the German Federal Ministry of Education and Research (BMBF) for this research project, as well as for the \enquote{OptimAgent Project}.

We would also like to express our sincere appreciation for the generous support provided by the German Research Foundation (DFG) within Research Training Group 2126 \enquote{Algorithmic Optimization}.

\printbibliography


\end{document}